%% file: paper.tex
\documentclass{article}
\pdfoutput=1
\usepackage{arxiv}
\usepackage{amsmath, amssymb, amsthm, bm}
\usepackage{mathrsfs}
\usepackage{hyperref, url}
\usepackage{subcaption}
\usepackage[shortlabels]{enumitem}
\usepackage{float, graphicx, placeins}
\graphicspath{{figures/}}
\usepackage{booktabs, multirow}
\usepackage{algorithm}
\usepackage[noend]{algpseudocode}
\usepackage{multirow}
\usepackage{multicol}
\usepackage{tikz}

\hypersetup{
    colorlinks=true,
    linkcolor=blue,
    filecolor=magenta,      
    urlcolor=cyan,
    pdftitle={The Price of Interpretability},
    pdfpagemode=FullScreen,
}

\theoremstyle{plain}
    \newtheorem{theorem}{Theorem}
    \newtheorem{proposition}{Proposition}

\theoremstyle{remark}
    \newtheorem*{remark}{Remark}
    
\theoremstyle{definition}
    \newtheorem{definition}{Definition}

\newcommand{\norm}[1]{\left\lVert#1\right\rVert}

\newcommand{\parenth}[1]{\left( #1 \right)}
\newcommand{\R}{\mathbb{R}}
\newcommand{\N}{\mathbb{N}}

\title{The Price of Interpretability}
\author{Dimitris Bertsimas, Arthur Delarue, Patrick Jaillet, Sebastien Martin}

\author{
    Dimitris Bersimas \\
    Sloan School of Management\\
    Massachusetts Institute of Technology\\
    Cambridge, MA 02139 \\
    \texttt{dbertsim@mit.edu} \\
    \And
    Arthur Delarue \\
    Operations Research Center\\
    Massachusetts Institute of Technology\\
    Cambridge, MA 02139 \\
    \texttt{adelarue@mit.edu} \\
    \And
    Patrick Jaillet \\
    Dep. of Electrical Engineering and Computer Science\\
    Massachusetts Institute of Technology\\
    Cambridge, MA 02139 \\
    \texttt{jaillet@mit.edu} \\
    \And
    Sebastien Martin \\
    Operations Research Center\\
    Massachusetts Institute of Technology\\
    Cambridge, MA 02139 \\
    \texttt{92sebastien@gmail.com} \\
}

\begin{document}
\maketitle

\begin{abstract}
    When quantitative models are used to support decision-making on complex and important topics, understanding a model's ``reasoning'' can increase trust in its predictions, expose hidden biases, or reduce vulnerability to adversarial attacks.
    However, the concept of interpretability remains loosely defined and application-specific.
    In this paper, we introduce a mathematical framework in which machine learning models are constructed in a sequence of \emph{interpretable steps}.
    We show that for a variety of models, a natural choice of interpretable steps recovers standard interpretability proxies (e.g., sparsity in linear models).
    We then generalize these proxies to yield a parametrized family of consistent measures of model interpretability.
    This formal definition allows us to quantify the ``price'' of interpretability, i.e., the tradeoff with predictive accuracy. We demonstrate practical algorithms to apply our framework on real and synthetic datasets.
\end{abstract}


\section{Introduction}\label{sec:intro}

\input{1-intro}

\section{A Sequential View of Model Construction}\label{sec:paths}

\input{2-paths}

\section{The Tradeoffs of Interpretability}\label{sec:metrics}

\input{3-metrics}

\section{Coherent Interpretability Losses}\label{sec:gammas}

\input{4-gammas}

\section{Interpretability Losses in Practice}\label{sec:computation}

\input{5-computation}

\section{Application: Linear Regression}\label{sec:linreg}

\input{6-linreg}

\section{Conclusions}

In this paper, we have presented a simple optimization-based framework to model the interpretability of machine learning models. 
Our framework provides a new way to think about what interpretability means to users in different applications and quantify how this meaning affects the tradeoff with predictive accuracy.
This framework is general, and each application can have its own modeling and optimization challenges, and could be an opportunity for further research.

\section*{Acknowledgements}
Research funded in part by ONR grant N00014-18-1-2122.

\FloatBarrier
\bibliographystyle{plain}
\bibliography{references}

\normalsize
\appendix
\section{Appendix}

\input{appendix}

\end{document}

%% file: 1-intro.tex

Predictive models are used in an increasingly high-stakes set of applications, from bail decisions in the criminal justice system \cite{Berk2017, Kleinberg2017} to treatment recommendations in personalized medicine \cite{Bertsimas2017}.
As the stakes have risen, so has the negative impact of incorrect predictions, which could be due to a poorly trained model or to undetected confounding patterns within the data itself \cite{Mullainathan2017}.

As machine learning models influence a growing fraction of everyday life, individuals often want to understand the reasons for the decisions that affect them.
Many governments now recognize a ``right to explanation'' for significant decisions, for instance as part of the European Union's General Data Protection Regulation \cite{Goodman2016}.
However, state-of-the-art machine learning methods such as random forests and neural networks are black boxes: their complex structure makes it difficult for humans, including domain experts, to understand their predictive behavior \cite{Breiman2001a, Friedman2001}.

\subsection{Interpretable Machine Learning}

According to Leo Breiman \cite{Breiman2001}, machine learning has two objectives: prediction, i.e., determining the value of the target variable for new inputs, and information, i.e., understanding the natural relationship between the input features and the target variable.
Studies have shown that many decision makers exhibit an inherent distrust of automated predictive models, even if they are proven to be more accurate than human forecasters \cite{Dietvorst2014}.
One way to overcome ``algorithm aversion'' is to give decision makers agency to modify the model's predictions \cite{Dietvorst2016}.
Another is to provide them with understanding.

Many studies in machine learning seek to train more interpretable models in lieu of complex black boxes.
Decision trees \cite{Breiman1984, Bertsimas2017b} are considered interpretable for their discrete structure and graphical visualization, as are close relatives including rule lists \cite{Letham2015, Yang2016a}, decision sets \cite{Lakkaraju2016}, and case-based reasoning \cite{Kim2015}.
Other approaches include generalized additive models \cite{Lou2012}, i.e., linear combinations of single-feature models, and score-based methods \cite{Ustun2016}, where integer point values for each feature can be summed up into a final ``score''.
In the case of linear models, interpretability often comes down to sparsity (small number of nonzero coefficients), a topic of extensive study over the past twenty years \cite{Hastie2015}.
Sparse regression models can be trained using heuristics such as LASSO, stagewise regression or least-angle regression \cite{Tibshirani1996,Taylor2015,Efron2004}, or scalable mixed-integer approaches \cite{Bertsimas2016, Bertsimas2017a}.

Many practitioners are hesitant to give up the high accuracy of black box models in the name of interpretability, and prefer to construct \emph{ex post} explanations for a model's predictions.
Some approaches create a separate explanation for each prediction in the dataset, e.g. by approximating the nonlinear decision boundary of a neural network with a hyperplane \cite{Ribeiro2016}.
Others define metrics of feature importance to quantify the effect of each feature in the overall model \cite{Friedman2001,Datta2016}.

Finally, some approaches seek to approximate a large, complex model such as a neural network or a random forest with a simpler one -- a decision tree \cite{Bastani2017}, two-level rule list \cite{Lakkaraju2017}, or smaller neural network \cite{Bucila2006}.
Such \emph{global} explanations can help human experts detect systemic biases or confounding variables.
However, even if these approximations are almost as accurate as the original model, they may have very different behavior on some inputs and can thus provide a misleading assessment of the model's behavior \cite{Gilpin2018}.

The interpretability of linear models and the resulting tradeoff with predictive accuracy are of significant interest to the machine learning community \cite{Freitas2014}.
However, a major challenge in this line of research is that the very concept of interpretability is hard to define and even harder to quantify \cite{Lipton2016}.
Many definitions of interpretability have a ``know it when you see it" aspect which impedes quantitative analysis: though some aspects of interpretability are easy to measure, others may be difficult to evaluate without human input \cite{Doshi-Velez2017}.

\subsection{Contributions}

We introduce the framework of \emph{interpretable paths}, in which models are decomposed into simple building blocks.
An interpretable path is a sequence of models of increasing complexity which can represent a sequential process of ``reading'' or ``explaining'' a model.
Using examples of several machine learning model classes, we show that the framework of interpretable paths is relevant and intuitively captures properties associated with interpretability.

To formalize which paths are more interpretable, we introduce path interpretability metrics.
We define \emph{coherence} conditions that such metrics should satisfy, and derive a parametric family of coherent metrics.

This study of interpretable paths naturally leads to a family of \emph{model} interpretability metrics.
The proposed metrics generalize a number of proxies for interpretability from the literature, such as sparsity in linear models and number of splits for decision trees, and also encompass other desirable characteristics.

The model interpretability metrics can be used to select models that are both accurate and interpretable.
To this end, we formulate the optimization problem of computing models that are on the Pareto front of interpretability and predictive accuracy (price of interpretability).
We give examples in various settings, and discuss computational challenges.

We study an in-depth application to linear models on real and synthetic datasets.
We discuss both the modeling aspect (the choice of the interpretability metric), as well as the computational aspect for which we propose exact mixed-integer formulations and scalable local improvement heuristics.

%% file: 2-paths.tex


\subsection{Selecting a Model}\label{sec:paths-space}


Most machine learning problems can be viewed through the lens of optimization. Given a set of models $\mathcal{M}$, each model $m\in\mathcal{M}$ is associated with a cost $c(m) > 0$, typically derived from data, representing the performance of the model on the task at hand (potentially including a regularization term).
Training a machine learning model means choosing the appropriate $m$ from $\mathcal{M}$ (for example the one that minimizes $c(m)$).
To make this perspective more concrete, we will use the following examples throughout the paper.

\paragraph{Linear models.}
Given the feature matrix ${X}\in\mathbb{R}^{n\times d}$ of a dataset of size $n$ with feature space in $\mathbb{R}^d$ and the corresponding vector of labels ${y}\in\mathbb{R}^n$, a linear model corresponds to a set of linear coefficients $\beta\in\mathbb{R}^d$.
In this example, $\mathcal{M}=\mathbb{R}^d$, and the cost $c(\cdot)$ depends on the application: for ordinary least squares (OLS), $c(\beta)=(1/n)\|X\beta-y\|^2$ (mean squared error).

\paragraph{Classification trees (CART).}
In this case, each model corresponds to a binary decision tree structure \cite{Breiman1984}, so $\mathcal{M}$ is the set of all possible tree structures of any size.
Given a tree $t\in\mathcal{M}$ and an input $x\in\mathbb{R}^d$, let $t(x)$ designate the tree's estimate of the corresponding label. Then a typical performance metric $c(t)$ is the number of misclassified points.
If we have a dataset with $n$ points $(x_1, \cdots, x_n) \in (\mathbb{R}^d)^n$ associated with classification labels $(y_1, \cdots, y_n) \in \{0,1\}^n$ then we have $c(t) = \sum_{i=1}^n \bm{1}(t(x_i) \neq y_i)$.
    
\paragraph{Clustering.}
We consider the k-means clustering problem for a dataset $\mathcal{D}$ of $n$ points in dimension $d$.
Our model space $\mathcal{M}$ is the set of all partitions of the dataset, each partition representing a cluster.
Formally $\mathcal{M} = \cup_{K=1}^n\{(A_1, \ldots, A_K): i\neq j\Rightarrow A_i\cap A_j=\emptyset, \cup_{i=1}^K A_i=\mathcal{D}\}$.
To evaluate a partition, we can use the within-cluster sum of squares $c(A_1, \ldots, A_K) = \sum_{k=1}^K \sum_{x_i\in A_k} \|x_i - \mu_k\|^2$, where $\mu_k=\sum_{x_i\in A_k}x_i/|A_k|$ is the centroid of cluster $A_k$.

\subsection{Interpretable Steps} \label{sec:steps}

For our guiding examples, typical proxies for interpretability include sparsity in linear models \cite{Bertsimas2016}, a small number of nodes in a classification tree \cite{Bertsimas2017b}, or a small number of clusters.

As we try to rationalize why these are good proxies for interpretability, one possible approach is to consider how humans read and explain these models.
For example, a linear model is typically introduced coefficient by coefficient, a tree is typically read node by node from the root to the leaves, and clusters are typically examined one by one.
During this process, we build a model that is more and more complex.
In other words, the human process of understanding a model can be viewed as a decomposition into simple building blocks.

We introduce the notion of an \emph{interpretable step} to formalize this sequential process.
For every model $m\in\mathcal{M}$, we define a step neighborhood function $\mathcal{S}$ that associates each model $m$ to the set of models $\mathcal{S}(m)\subseteq\mathcal{M}$ such that $m'$ is one interpretable step away from $m$ if and only if $m' \in \mathcal{S}(m)$.
Interpretable steps represent simple model updates that can be chained to build increasingly complex models.

\begin{figure}[h]
    \centering
    \includegraphics[width=0.6\columnwidth]{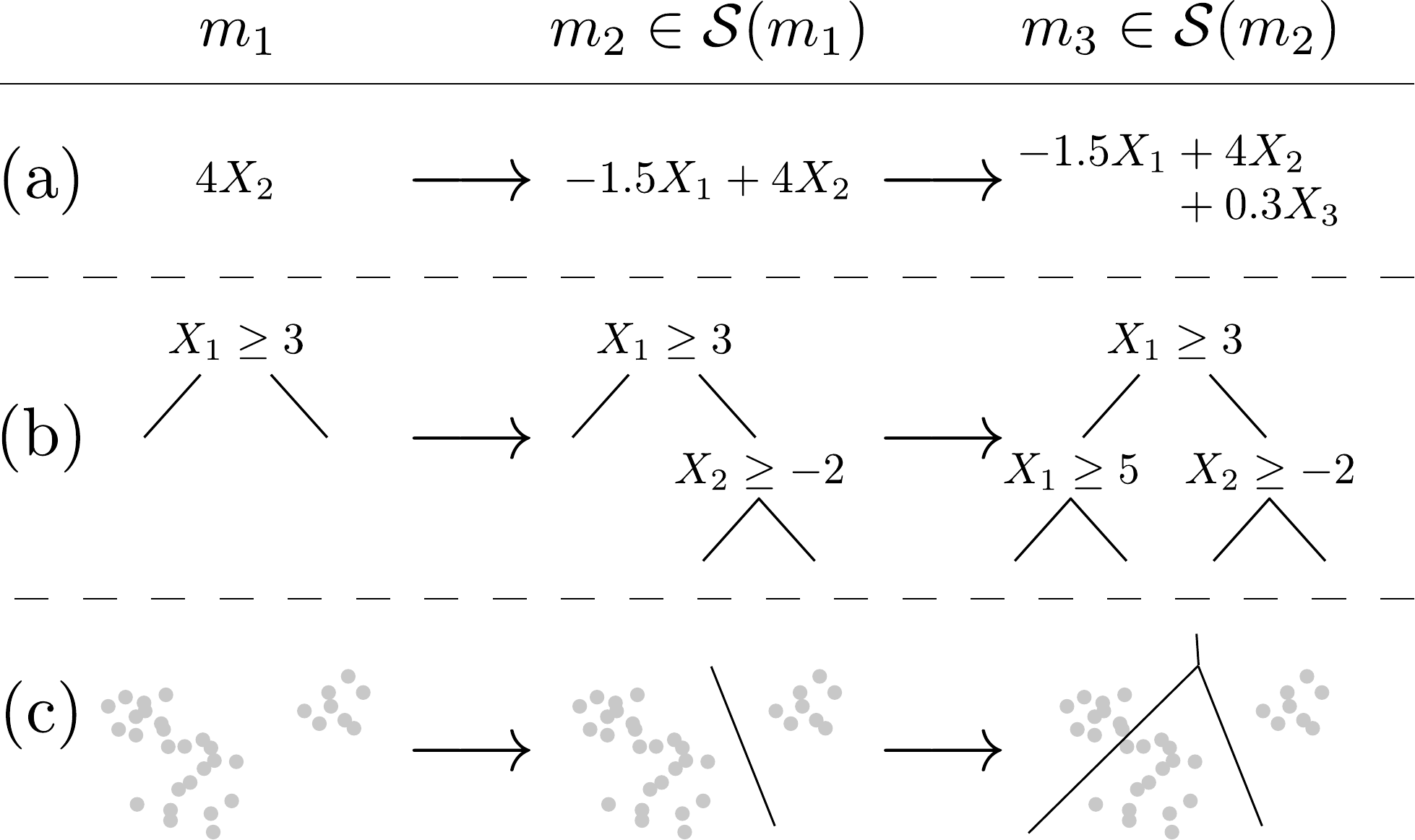}
    \caption{Illustration of the interpretable path framework with the three examples introduced in Section~\ref{sec:paths-space}:
    (a) is our linear model setting;
    (b) corresponds to the classification trees (CART);
    (c) to the clustering setting (in 2 dimensions).
    For each space of model, we illustrate an example of interpretable path following the choice of steps introduced in Section~\ref{sec:steps}.
    Each path has 3 steps: $m_1$, $m_2$ and $m_3$.}
    \label{fig:models-stepbystepmodels}
\end{figure}

For linear models, one possible interpretable step is modifying a single coefficient, i.e. $\beta'$ belongs to $\mathcal{S}(\beta)$ if $\norm{\beta-\beta'}_0\le 1$ ($\beta$ and $\beta'$ differ in at most one coefficient).
For CART, an interpretable step could be adding a split to an existing tree, i.e., $t' \in \mathcal{S}(t)$ if $t'$ can be obtained by splitting a leaf node of $t$ into two leaves.
For clustering, we could use the structure of hierarchical clustering and choose a step that increases the number of clusters by one by splitting an existing cluster into two. These examples are illustrated in Figure~\ref{fig:models-stepbystepmodels}.

Choosing the step neighborhood function $\mathcal{S}$ is a modeling choice and for the examples considered, there may be many other ways to define it.
To simplify the analysis, we only impose that $\mathcal{S}(m)\neq \emptyset$ for all $m$ (there must always be a feasible next step from any model), which can trivially be satisfied by ensuring $m\in\mathcal{S}(m)$ (an interpretable step can involve no changes to the model).

Given the choice of an interpretable step $\mathcal{S}$, we can define an \emph{interpretable path} of length $K$ as a sequence of $K$ models $\bm{m} = (m_1, \cdots, m_K)$ such that $m_k \in \mathcal{S}(m_{k-1})$ for all $1\le k \le K$, i.e., a sequence of interpretable steps starting from a base model $m_0$.
The choice of $m_0$, the ``simplest'' model, is usually obvious: in our examples, $m_0$ could be a linear model with $\beta= \bf{0}$, an empty classification tree, or a single cluster containing all data points.
Given the model space $\mathcal{M}$, we call $\mathcal{P}_K$ the set of all interpretable paths of length $K$ and $\mathcal{P}=\cup_{K=0}^{\infty}\mathcal{P}_K$ the set of all interpretable paths of any length. 

Let us consider an example with classification trees to build intuition about interpretable paths.
The \texttt{iris} dataset is a small dataset often used to illustrate classification problems.
It records the petal length and width and sepal length and width of various iris flowers, along with their species (setosa, versicolor and virginica).
For simplicity, we only consider two of the four features (petal length and width) and subsample 50 total points from two of the three classes (versicolor and virginica).

\begin{figure}[h]
    \centering
    \begin{subfigure}[t]{0.33\columnwidth}
    \centering
    \begin{tikzpicture}
    \centering
    \footnotesize
    \node [above] at (0.0, 1.5) {width $\le 0.37$};
    \draw (0.0, 1.5) -- (1.0, 1.0);
    \draw (0.0, 1.5) -- (-1.0, 1.0);
    \node [right] at (0.5, 1.3) {{\tiny F}};
    \node [left] at (-0.5, 1.3) {{\tiny T}};
    \node [below] at (1.0, 1.0) {\textcolor{blue}{virginica}};
    \node [below] at (-1.0, 1.0) {\textcolor{red}{versicolor}};
    \draw [color=white] (1.0, 0.5) -- (2.0, 0.2);
    \draw [color=white] (1.0, 0.5) -- (0.67, 0.2);
    \draw [color=white] (-1.0, 0.5) -- (-2.0, 0.2);
    \draw [color=white] (-1.0, 0.5) -- (-0.67, 0.2);
    \node [below] at (2.0, 0.2) {\textcolor{white}{virginica\phantom{l}}};
    \node [below] at (0.67, 0.2) {\textcolor{white}{versicolor\phantom{g}}};
    \node [below] at (-2.0, 0.2) {\textcolor{white}{versicolor\phantom{g}}};
    \node [below] at (-0.67, 0.2) {\textcolor{white}{virginica\phantom{l}}};
    \end{tikzpicture}
    \includegraphics[width=\columnwidth]{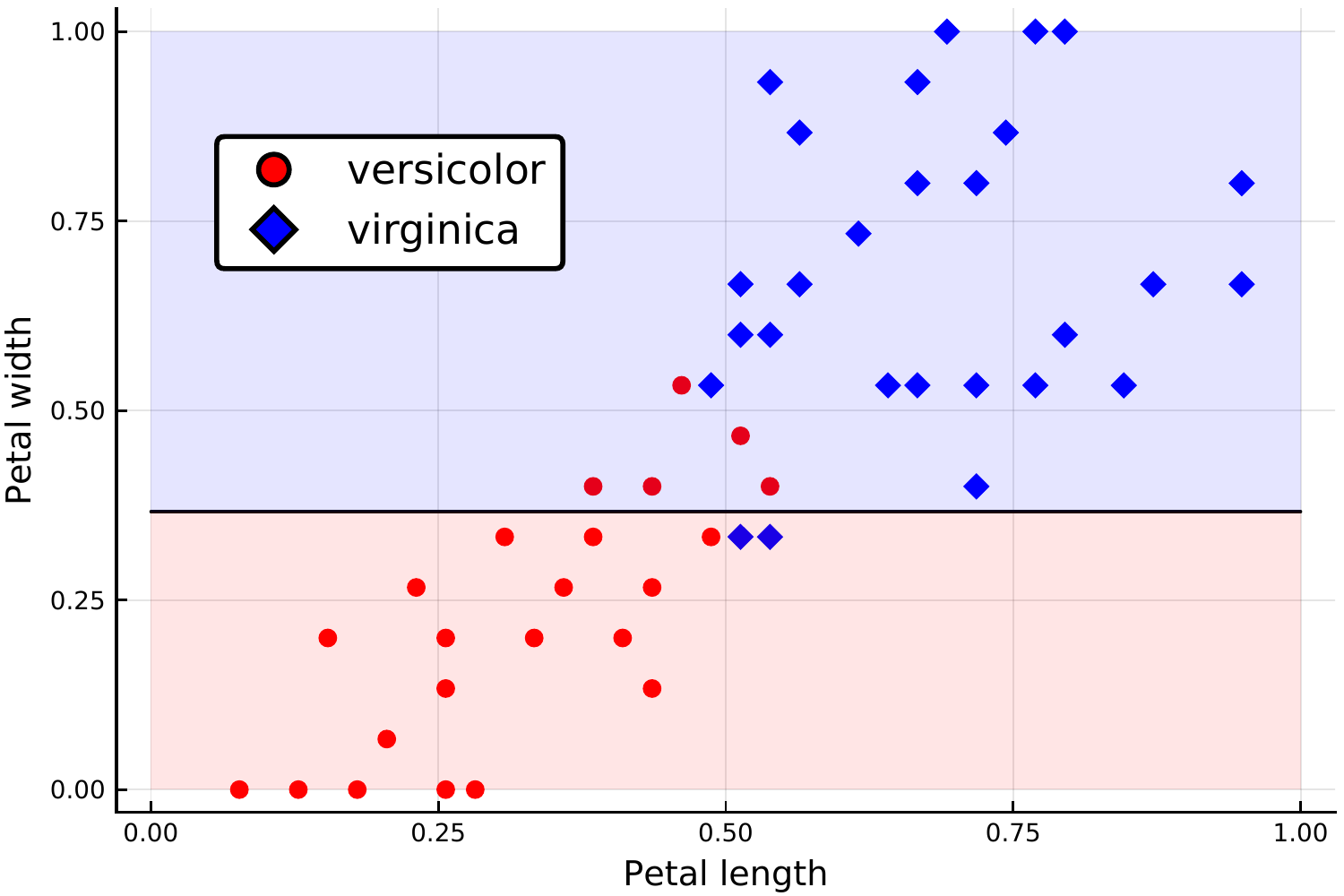}
    \caption{7 misclassified points}
    \end{subfigure}%
    \begin{subfigure}[t]{0.33\columnwidth}
    \centering
    \begin{tikzpicture}
    \centering
    \footnotesize
    \node [above] at (0.0, 1.5) {width $\le 0.37$};
    \draw (0.0, 1.5) -- (1.0, 1.0);
    \draw (0.0, 1.5) -- (-1.0, 1.0);
    \node [right] at (0.5, 1.3) {{\tiny F}};
    \node [left] at (-0.5, 1.3) {{\tiny T}};
    \node [below] at (1.0, 1.0) {width $\le 0.5$};
    \node [below] at (-1.0, 1.0) {\textcolor{red}{versicolor}};
    \draw (1.0, 0.5) -- (2.0, 0.2);
    \draw (1.0, 0.5) -- (0.67, 0.2);
    \draw [color=white] (-1.0, 0.5) -- (-2.0, 0.2);
    \draw [color=white] (-1.0, 0.5) -- (-0.67, 0.2);
    \node [below] at (2.0, 0.2) {\textcolor{blue}{virginica\phantom{l}}};
    \node [below] at (0.67, 0.2) {\textcolor{red}{versicolor\phantom{g}}};
    \node [below] at (-2.0, 0.2) {\textcolor{white}{versicolor\phantom{g}}};
    \node [below] at (-0.67, 0.2) {\textcolor{white}{virginica\phantom{l}}};
    \end{tikzpicture}
    \includegraphics[width=\columnwidth]{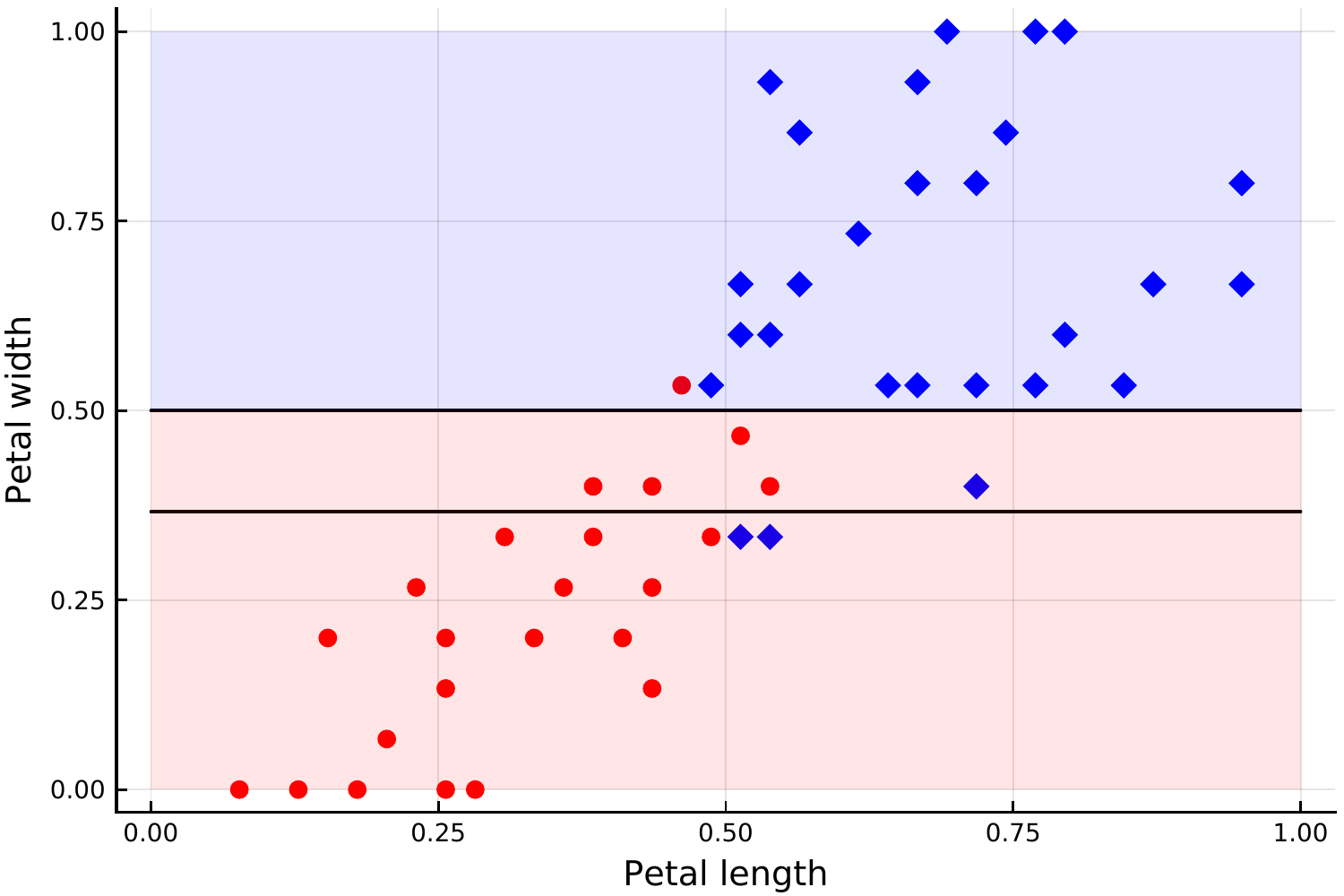}
    \caption{4 misclassified points}
    \end{subfigure}%
    \begin{subfigure}[t]{0.33\columnwidth}
    \centering
    \begin{tikzpicture}
    \centering
    \footnotesize
    \node [above] at (0.0, 1.5) {width $\le 0.37$};
    \draw (0.0, 1.5) -- (1.0, 1.0);
    \draw (0.0, 1.5) -- (-1.0, 1.0);
    \node [right] at (0.5, 1.3) {{\tiny F}};
    \node [left] at (-0.5, 1.3) {{\tiny T}};
    \node [below] at (1.0, 1.0) {width $\le 0.5$};
    \node [below] at (-1.0, 1.0) {length $\le 0.5$};
    \draw (1.0, 0.5) -- (2.0, 0.2);
    \draw (1.0, 0.5) -- (0.67, 0.2);
    \draw (-1.0, 0.5) -- (-2.0, 0.2);
    \draw (-1.0, 0.5) -- (-0.67, 0.2);
    \node [below] at (2.0, 0.2) {\textcolor{blue}{virginica\phantom{l}}};
    \node [below] at (0.67, 0.2) {\textcolor{red}{versicolor\phantom{g}}};
    \node [below] at (-2.0, 0.2) {\textcolor{red}{versicolor\phantom{g}}};
    \node [below] at (-0.67, 0.2) {\textcolor{blue}{virginica\phantom{l}}};
    \end{tikzpicture}
    \includegraphics[width=\columnwidth]{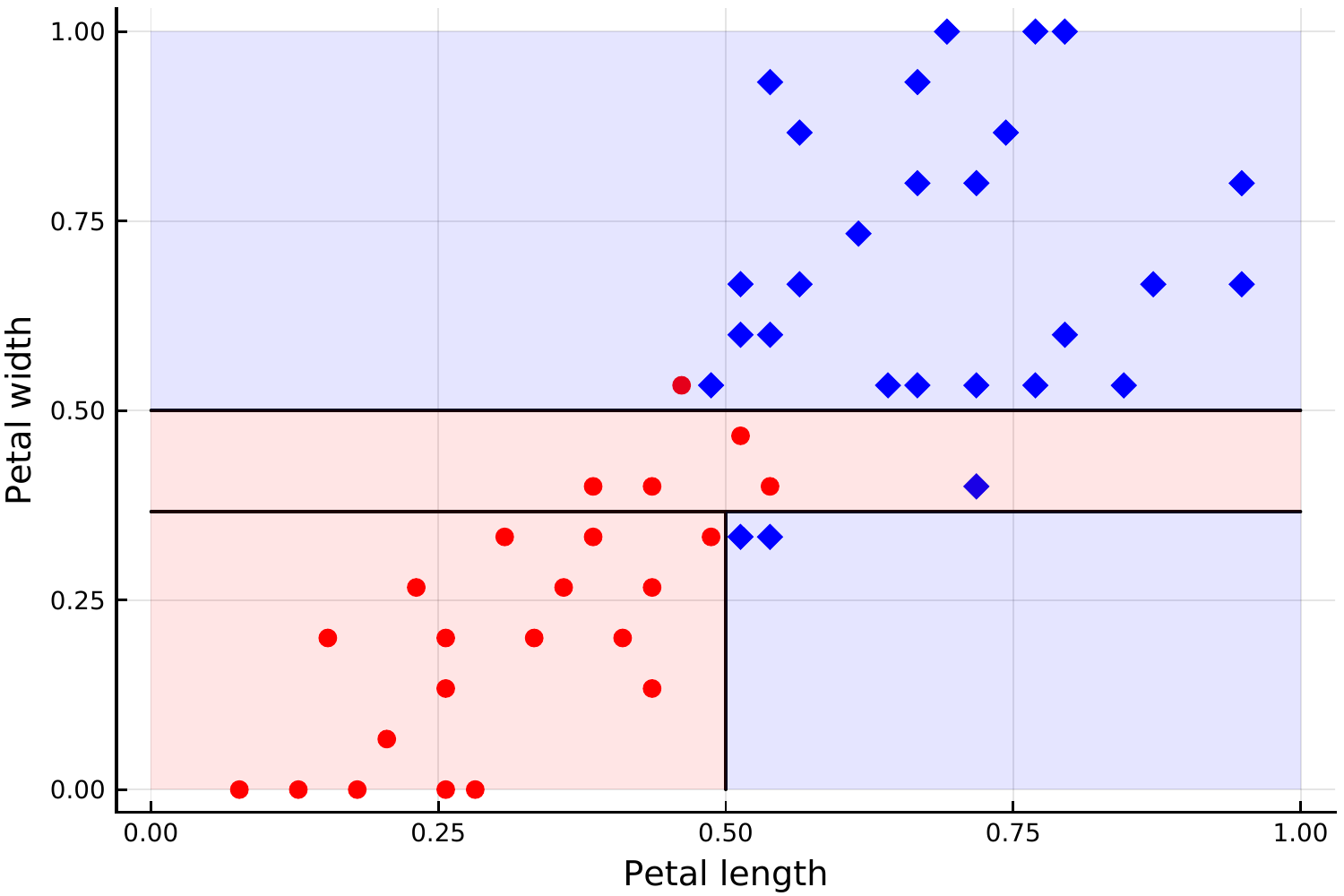}
    \caption{2 misclassified points}
    \end{subfigure}
    \caption{Visualization of an interpretable path leading to $t_{\text{bad}}$ (shown on the right).}
    \label{fig:badtree}
    \end{figure}
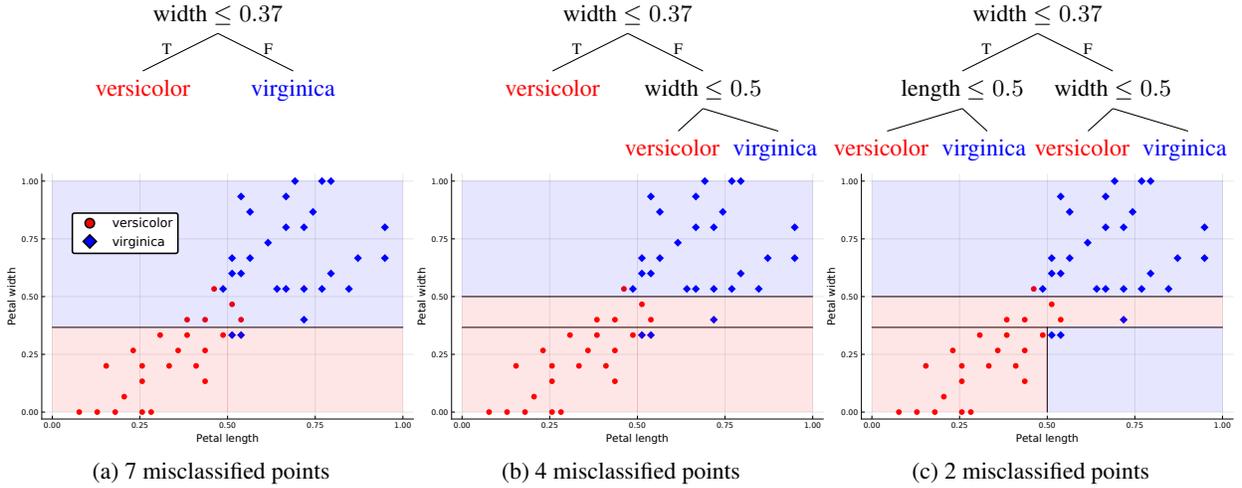
    
    \begin{figure}[h]
    \centering
    \begin{subfigure}[t]{0.33\columnwidth}
    \centering
    \begin{tikzpicture}
    \centering
    \footnotesize
    \node [above] at (0.0, 1.5) {width $\le 0.5$};
    \draw (0.0, 1.5) -- (1.0, 1.0);
    \draw (0.0, 1.5) -- (-1.0, 1.0);
    \node [right] at (0.5, 1.3) {{\tiny F}};
    \node [left] at (-0.5, 1.3) {{\tiny T}};
    \node [below] at (1.0, 1.0) {\textcolor{blue}{virginica\phantom{l}}};
    \node [below] at (-1.0, 1.0) {\textcolor{red}{versicolor\phantom{g}}};
    \draw [color=white] (1.0, 0.5) -- (2.0, 0.2);
    \draw [color=white] (1.0, 0.5) -- (0.67, 0.2);
    \draw [color=white] (-1.0, 0.5) -- (-2.0, 0.2);
    \draw [color=white] (-1.0, 0.5) -- (-0.67, 0.2);
    \node [below] at (2.0, 0.2) {\textcolor{white}{virginica\phantom{l}}};
    \node [below] at (0.67, 0.2) {\textcolor{white}{versicolor\phantom{g}}};
    \node [below] at (-2.0, 0.2) {\textcolor{white}{versicolor\phantom{g}}};
    \node [below] at (-0.67, 0.2) {\textcolor{white}{virginica\phantom{l}}};
    \end{tikzpicture}
    \includegraphics[width=\columnwidth]{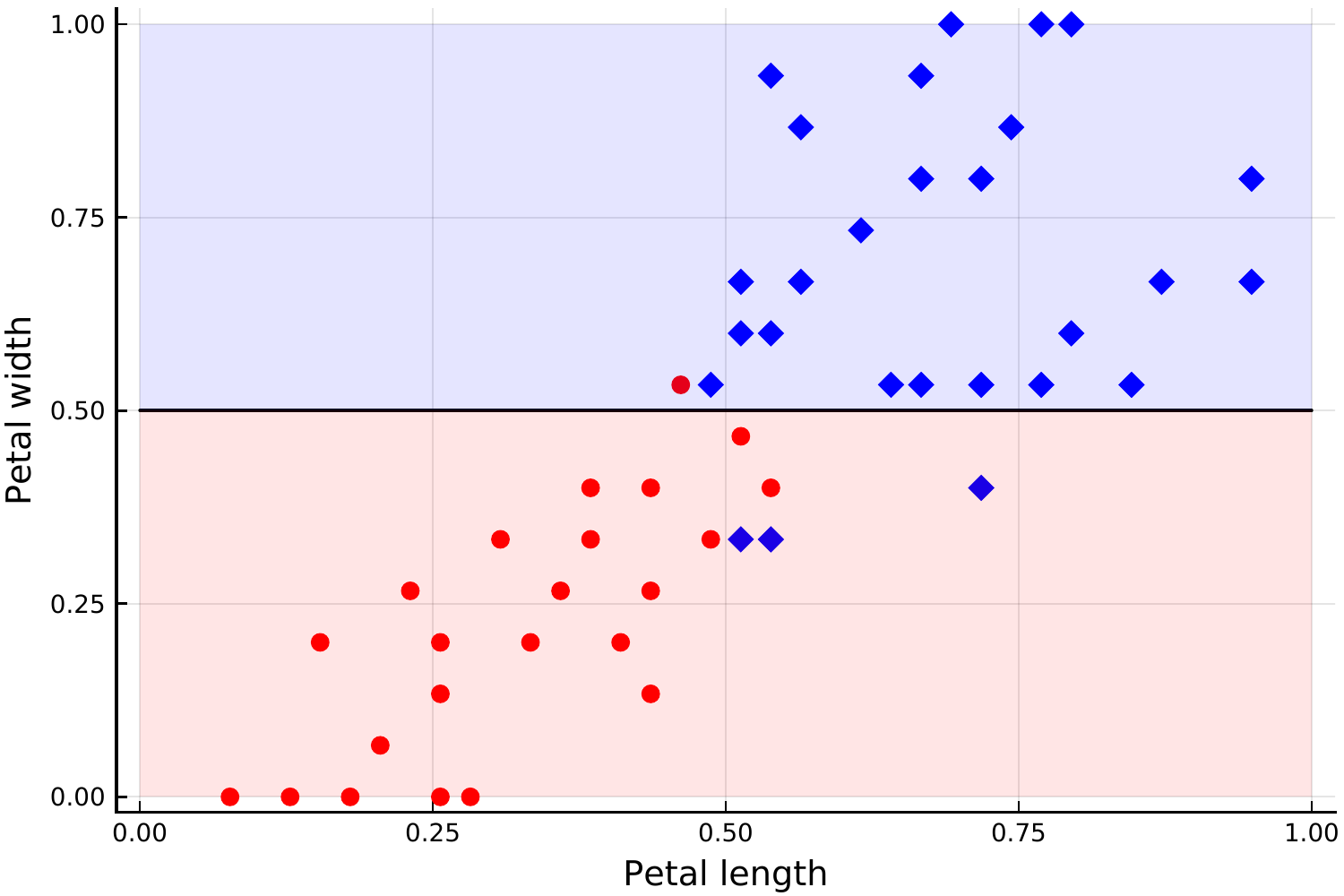}
    \caption{4 misclassified points}
    \end{subfigure}%
    \begin{subfigure}[t]{0.33\columnwidth}
    \centering
    \begin{tikzpicture}
    \centering
    \footnotesize
    \node [above] at (0.0, 1.5) {width $\le 0.5$};
    \draw (0.0, 1.5) -- (1.0, 1.0);
    \draw (0.0, 1.5) -- (-1.0, 1.0);
    \node [right] at (0.5, 1.3) {{\tiny F}};
    \node [left] at (-0.5, 1.3) {{\tiny T}};
    \node [below] at (1.0, 1.0) {\textcolor{blue}{virginica\phantom{l}}};
    \node [below] at (-1.0, 1.0) {length $\le 0.5$};
    \draw [color=white] (1.0, 0.5) -- (2.0, 0.2);
    \draw [color=white] (1.0, 0.5) -- (0.67, 0.2);
    \draw (-1.0, 0.5) -- (-2.0, 0.2);
    \draw (-1.0, 0.5) -- (-0.67, 0.2);
    \node [below] at (2.0, 0.2) {\textcolor{white}{virginica\phantom{l}}};
    \node [below] at (0.67, 0.2) {\textcolor{white}{versicolor\phantom{g}}};
    \node [below] at (-2.0, 0.2) {\textcolor{red}{versicolor\phantom{g}}};
    \node [below] at (-0.67, 0.2) {\textcolor{blue}{virginica\phantom{l}}};
    \end{tikzpicture}
    \includegraphics[width=\columnwidth]{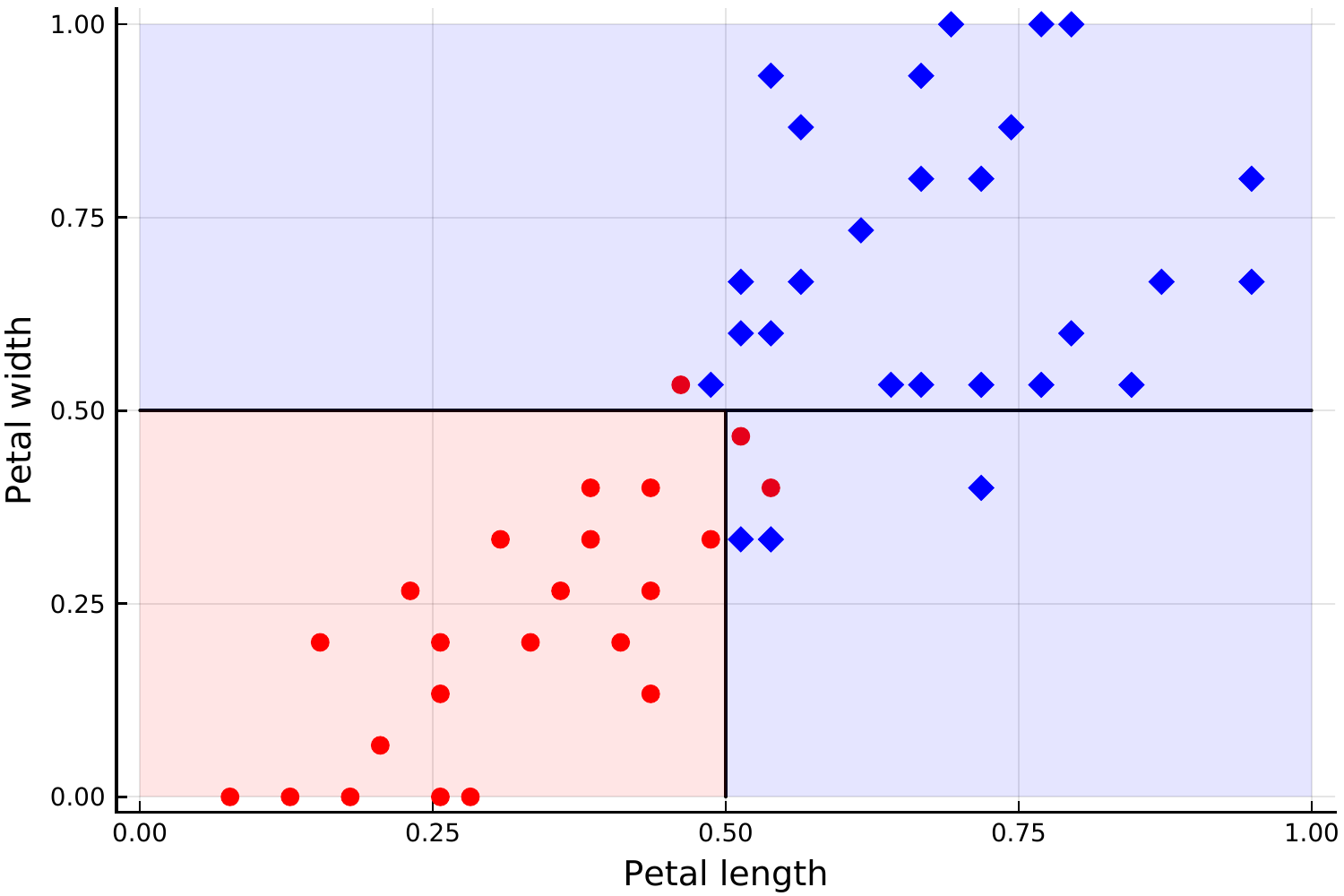}
    \caption{3 misclassified points}
    \end{subfigure}%
    \begin{subfigure}[t]{0.33\columnwidth}
    \centering
    \begin{tikzpicture}
    \centering
    \footnotesize
    \node [above] at (0.0, 1.5) {width $\le 0.5$};
    \draw (0.0, 1.5) -- (1.0, 1.0);
    \draw (0.0, 1.5) -- (-1.0, 1.0);
    \node [right] at (0.5, 1.3) {{\tiny F}};
    \node [left] at (-0.5, 1.3) {{\tiny T}};
    \node [below] at (1.0, 1.0) {length $\le 0.47$};
    \node [below] at (-1.0, 1.0) {length $\le 0.5$};
    \draw (1.0, 0.5) -- (2.0, 0.2);
    \draw (1.0, 0.5) -- (0.67, 0.2);
    \draw (-1.0, 0.5) -- (-2.0, 0.2);
    \draw (-1.0, 0.5) -- (-0.67, 0.2);
    \node [below] at (2.0, 0.2) {\textcolor{blue}{virginica\phantom{l}}};
    \node [below] at (0.67, 0.2) {\textcolor{red}{versicolor\phantom{g}}};
    \node [below] at (-2.0, 0.2) {\textcolor{red}{versicolor\phantom{g}}};
    \node [below] at (-0.67, 0.2) {\textcolor{blue}{virginica\phantom{l}}};
    \end{tikzpicture}
    \includegraphics[width=\columnwidth]{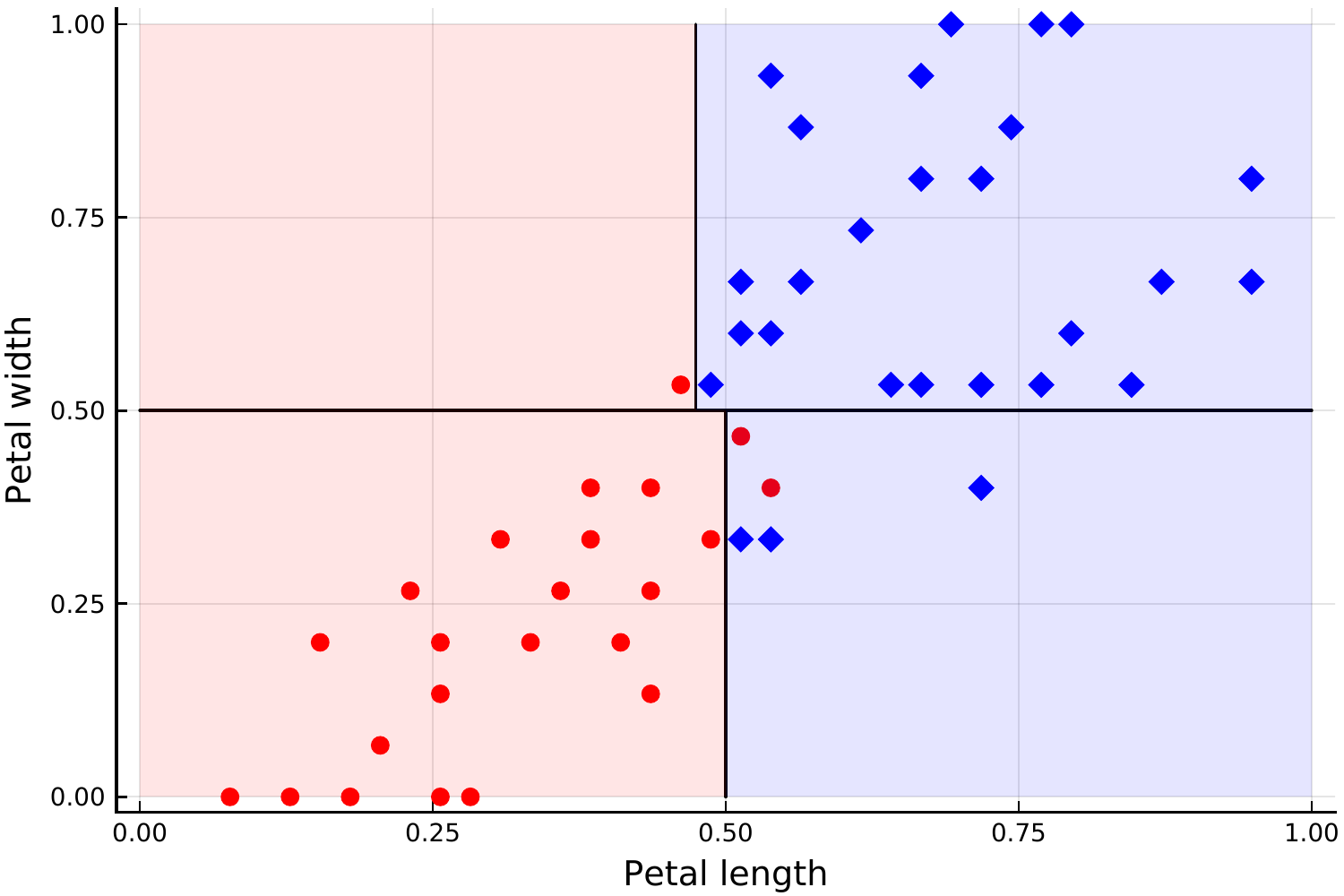}
    \caption{2 misclassified points}
    \end{subfigure}
    \caption{Visualization of an interpretable path leading to $t_{\text{good}}$ (shown on the right).}
    \label{fig:goodtree}
\end{figure}

We define an interpretable step as splitting one leaf node into two.
Given the \texttt{iris} dataset, we consider two classification trees $t_{\text{good}}, t_{\text{bad}} \in \mathcal{M}$.
Both trees have a depth of 2, exactly 3 splits, and a misclassification cost of 2.
However, when we consider interpretable paths leading to these two trees, we notice some differences.
An interpretable path $\bm{t_{\textbf{bad}}} \in \mathcal{P}$ leading to $t_{\text{bad}} \in \mathcal{M}$ is shown in Figure~\ref{fig:badtree}, and an interpretable path $\bm{t_{\textbf{good}}}$ leading to $t_{\text{good}}$ is detailed in Figure~\ref{fig:goodtree}.

For $t_{\text{bad}}$, the first split results in an intermediate tree with a high classification error, which could be less intuitive for flower connoisseurs.
In contrast, the first split of $t_{\text{good}}$ gives a much more accurate intermediate tree. We will introduce a way to formally identify which of the two paths is more interpretable.

%% file: 3-metrics.tex

In this section, we consider the choice of an interpretability loss $\mathcal{L}(\bm{m})$ for all interpretable paths $\bm{m}\in\mathcal{P}$ such that a path $\bm{m}$ is considered more interpretable than a path $\bm{m'}$ if and only if $\mathcal{L}(\bm{m})<\mathcal{L}(\bm{m'})$. 
We first motivate this formalism by showing it can lead to a notion of interpretability loss for models as well.
We then use a simple example to build intuition about the choice of $\mathcal{L}$. 
\subsection{From paths to models} \label{sec:paths-complexity}

Defining a loss function for the interpretability of a path can naturally lead to an interpretability loss on the space of models, with the simple idea that more interpretable paths should lead to more interpretable models.
Given a path interpretability loss $\mathcal{L}(\cdot)$, we can define a corresponding model interpretability loss as
\begin{equation}
\mathcal{L}(m) = \begin{cases}
            \infty, &\text{if } \mathcal{P}(m) = \emptyset,\\
            \min_{\bm{m} \in \mathcal{P}(m)} \mathcal{L}(\bm{m}), &\text{otherwise,}
        \end{cases}
\label{eq:path-to-model-interpretability}
\end{equation}
where $\mathcal{P}_K(m)=\{\bm{m}\in\mathcal{P}_K, m_K=m\}$ designates the set of interpretable paths of length $K$ leading to $m$, and $\mathcal{P}(m)=\cup_{K=0}^{\infty}\mathcal{P}_K(m)$ designates the set of finite interpretable paths leading to $m$.
In other words, the interpretability loss of a model $m$ is the interpretability loss of the most interpretable path leading to $m$.

As an example, consider the following path interpretability loss, which we call \emph{path complexity} and define as $\mathcal{L}_{\mathrm{complexity}}(\bm{m}) = |\bm{m}|$ (number of steps in the path). Under this metric, paths are considered less interpretable if they are longer.
From~\eqref{eq:path-to-model-interpretability} we can then define the interpretability loss of a given model \(m\) as
\[
\mathcal{L}_{\mathrm{complexity}}(m) =  \min_{\bm{m} \in \mathcal{P}(m)} |\bm{m}|,
\]
which corresponds to the minimal number of interpretable steps required to reach $m$.

In the context of the examples from Section~\ref{sec:paths}, the function $\mathcal{L}_{\text{complexity}}$ recovers typical interpretability proxies.
For a linear model $\beta$, \(\mathcal{L}_{\mathrm{complexity}}(\beta) = \|\beta\|_0\) is the sparsity of the model (number of non-zero coefficients).
For a classification tree $t$, \(\mathcal{L}_{\mathrm{complexity}}(t)\) is the number of splits.
In a clustering context, \(\mathcal{L}_{\mathrm{complexity}}(A_1,\cdots,A_K) = K\) is just the number of clusters.
We refer to this candidate loss function as the \emph{model complexity}.

A fundamental problem of interpretable machine learning is finding the highest-performing model at a given level of interpretability \cite{Breiman2001}.
Defining an interpretability loss function $\mathcal{L}(\cdot)$ on the space of models $\mathcal{M}$ is important because it allows us to formulate this problem generally as follows:
\begin{equation}
\min_{m\in\mathcal{M}}c(m)\quad\text{s.t.}\quad \mathcal{L}(m)\le \ell,
\label{eq:path-fundamentalproblem}
\end{equation}
where $\ell$ is the desired level of interpretability.
Problem~\eqref{eq:path-fundamentalproblem} produces models on the Pareto front of accuracy and interpretability.
If we compute this Pareto front by solving problem~\eqref{eq:path-fundamentalproblem} for any $\ell$, then we can mathematically characterize the \emph{price} of our definition of interpretability on our dataset given a class of models, making the choice of a final model easier.

In the case of model complexity $\mathcal{L}_{\text{complexity}}$, for $\ell=K$ problem~\eqref{eq:path-fundamentalproblem} can be written as
\begin{equation} \label{eq:model-mincomplexity}
    \min_{\bm{m} \in \mathcal{P}_K}  \quad c(m_K).
\end{equation}

Problem~\eqref{eq:model-mincomplexity} generalizes existing problems in interpretable machine learning: best subset selection (\(L_0\)-constrained sparse regression) for linear models \cite{Bertsimas2016}, finding the best classification tree of a given size \cite{Bertsimas2017b}, or the K-means problem of finding the $K$ best possible clusters.

Thus, the framework of interpretable paths naturally gives rise to a general definition of model complexity via the loss function $\mathcal{L}_{\mathrm{complexity}}$, and our model generalizes many existing approaches.
By some counts, however, model complexity remains an incomplete interpretability loss.
For instance, it does not differentiate between the trees $t_{\text{good}}$ and $t_{\text{bad}}$: both models have a complexity of 3 because they can be reached in three steps.
More generally, $\mathcal{L}_{\text{complexity}}$ does not differentiate between paths of the same length, or between models that can be reached by paths of the same length.

\subsection{Incrementality} \label{sec:paths-tradeoff}

In the decision tree example from Figures~\ref{fig:badtree} and \ref{fig:goodtree}, we observed that the intermediate trees leading to $t_{\text{good}}$ were more accurate than the intermediate trees leading to $t_{\text{bad}}$.
Evaluating the costs $c(m_k)$ of intermediate models along a path $\bm{m}$ may provide clues as to the interpretability of the final model.

Consider the following toy example, where the goal is to estimate a child's age \(y_\text{Age}\) given height \(X_\text{Height}\) and weight \(X_\text{Weight}\).
The normalized features \(X_\text{Height}\) and \(X_\text{Weight}\) have correlation \(\rho = 0.9\) and are both positively correlated with the objective.
Solving the OLS problem yields $\beta^* = (2.12, -0.94)$, i.e.,
\begin{equation}\label{eq:model-linearexample}
    y_\text{Age} = 2.12 \cdot X_\text{Height} - 0.94 \cdot X_\text{Weight} + \varepsilon,
\end{equation}
with \(\varepsilon =  X\beta^* - y\) the error term.
The mean squared error (MSE) of the model $\beta^*$ is \(c(\beta^*) = \frac{1}{n}\sum_i \varepsilon_i^2= 0.25\). 

As in Section~\ref{sec:paths}, we define an interpretable step to be modifying a single coefficient in the linear model, keeping all other coefficients constant.
In this case, consider the three interpretable paths in Table~\ref{tab:three-decompositions}. When using the complexity loss \(\mathcal{L}_{\mathrm{complexity}}\), the first two paths in the table are considered equally interpretable because they have the same length.
But are they?
Both verify $c(m_2)=c(m'_2)=0.25$, but $c(m_1)=1.13<c(m'_1)=4.74$.
Indeed, \(m'_1\) is a particularly inaccurate model, as weight is positively correlated with age.
And furthermore, if having an accurate first step matters to the user, then path $\bm{\hat{m}}$ may be preferred even though it is longer.

\begin{table}[h!]
    \centering
    \begin{subtable}[c]{0.33\linewidth}
    \centering
    \begin{tabular}{lr}
    \toprule
    $\bm{m}$ & $c(m_i)$\\
    \midrule
    $m_0=(0,0)$ & $2.04$\\
    $m_1=(2.12,0)$ & $1.13$\\
    $m_2=(2.12,-0.94)$ & $0.25$\\
    \bottomrule
    \end{tabular}
    \caption{Two steps, starting with height}
    \label{tab:decomposition1}
    \end{subtable}\hfill%
    \begin{subtable}[c]{0.33\linewidth}
    \centering
    \begin{tabular}{lr}
    \toprule
    $\bm{m'}$ & $c(m'_i)$\\
    \midrule
    $m'_0=(0,0)$ & $2.04$\\
    $m'_1=(0,-0.94)$ & $4.74$\\
    $m'_2=(2.12,-0.94)$ & $0.25$\\
    \bottomrule
    \end{tabular}
    \caption{Two steps, starting with weight}
    \label{tab:decomposition2}
    \end{subtable}\hfill%
    \begin{subtable}[c]{0.33\linewidth}
    \centering
    \begin{tabular}{lr}
    \toprule
    $\bm{\hat m}$ & $c(\hat m_i)$\\
    \midrule
    $\hat{m}_0=(0,0)$ & $2.04$\\
    $\hat{m}_1=(1.70,0)$ & $0.60$\\
    $\hat{m}_2=(1.70,-0.94)$ & $0.43$\\
    $\hat{m}_3=(2.12,-0.94)$ & $0.25$\\
    \bottomrule
    \end{tabular}
    \caption{Three steps}
    \label{tab:decomposition3}
    \end{subtable}
    \caption{Three decompositions of $m^*$ into a sequence of interpretable steps.}
    \label{tab:three-decompositions}
\end{table}

As discussed in Section~\ref{sec:paths}, an interpretable path $\bm{m}$ leading to model $m$ can be viewed as a decomposition of $m$ into a sequence of easily understandable steps.
The costs of intermediate models should play a role in quantifying the interpretability loss of a path; higher costs should be penalized, as we want to avoid nonsensical intermediate models such as $m'_1$.

One way to ensure that every step of an interpretable path adds value is a greedy approach, where the next model at each step is chosen by minimizing the cost $c(\cdot)$:
\begin{equation} \label{eq:model-greedy}
    m^{\mathrm{greedy}}_{k+1} \in \arg\min \left\{ c(m),\,  m \in \mathcal{S}(m^{\mathrm{greedy}}_k) \right\} \quad \forall k \geq 1.
\end{equation}

In our toy example, restricting ourselves to paths of length 2, this means selecting the best possible $m^{\mathrm{greedy}}_1$, and then the best possible $m^{\mathrm{greedy}}_2$ given $m^{\mathrm{greedy}}_1$, as in stagewise regression \cite{Taylor2015}.
This will not yield the best possible model achievable in two steps as in \eqref{eq:model-mincomplexity}, but the first step is guaranteed to be the best one possible.
Notice that $c(m^{\mathrm{greedy}}_1) = 0.42 < 1.13=c(m_1)$, but $c(m^{\mathrm{greedy}}_2) = 0.39 > 0.25=c(m_2)$. The improvement of the first model comes at the expense of the second step.

\begin{figure}[h]
    \centering
    \includegraphics[width=0.35\columnwidth]{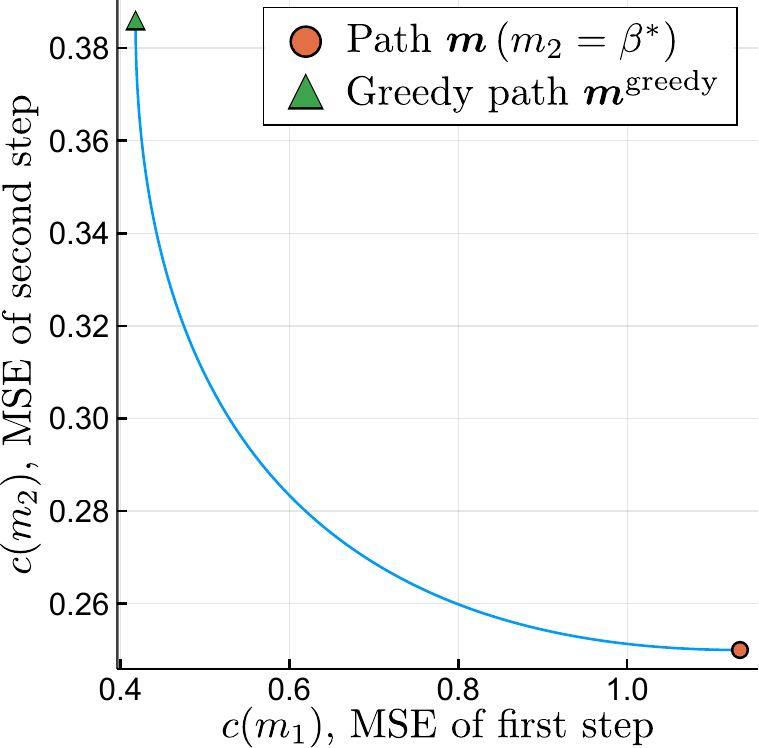}
    \caption{Tradeoff between the cost of the first and second models of the interpretable path.}
    \label{fig:models-2stepsgammatradeoff}
\end{figure}

Deciding which of the two paths $\bm{m}$ and $\bm{m}^\mathrm{greedy}$ is more interpretable is a hard question.
It highlights the tradeoff between the desirable incrementality of the greedy approach and the cost of the final model. For paths of length 2, there is a continuum of models between $\bm{m}$ and $\bm{m}^{\text{greedy}}$, corresponding to the Pareto front between $c(m_1)$ and $c(m_2)$, shown in Figure~\ref{fig:models-2stepsgammatradeoff}.

%% file: 4-gammas.tex

In the previous section, we developed intuition regarding the interpretability of different paths.
We now formalize this intuition in order to define a suitable interpretability loss.

\subsection{Coherent Path Interpretability Losses}

According to the loss $\mathcal{L}_{\text{complexity}}$ defined in Section~\ref{sec:paths-complexity}, which generalizes many notions of interpretability from the literature, a path is more interpretable if it is shorter.
In Section~\ref{sec:paths-tradeoff}, we saw that the cost of individual models along the path matters as well.

Sometimes, comparing the costs of intermediate models between two paths is easy because the cost of each step along one path is at least as good as the cost of the corresponding step in the other path.
In Table~\ref{tab:three-decompositions}, it is reasonable to consider $\bm{m}$ more interpretable than $\bm{m'}$ because $c(m_1)<c(m'_1)$ and $c(m_2)=c(m'_2)$.
In contrast, comparing the interpretability of $\bm{m}$ and $\bm{m}^{\text{greedy}}$ is more difficult and user-specific, because $c(m_1)>c(m_1^{\text{greedy}})$, but $c(m_2)<c(m_2^{\text{greedy}})$.

We now formalize this intuition into desirable properties of interpretability loss functions.
We first introduce the notion of a cost sequence, which provides a concise way to refer to the costs of all the steps in an interpretable path.
We then propose axioms for \emph{coherent} interpretability losses.

\begin{definition}[Cost sequence]\label{def:costsequence}
Given an interpretable path of length $K$, denoted as $\bm{m} \in \mathcal{P}_K$, the cost sequence $\bm{c}(\bm{m}) \in \R^\N$ is the infinite sequence $(c_1, c_2, \cdots)$ such that:
        \begin{equation*}
            c_k = \begin{cases}
                c(m_k), &\text{if } k \leq K,\\
                0, &\text{otherwise.}
            \end{cases}
        \end{equation*}
\end{definition}

\begin{definition}[Coherent Interpretability Loss]\label{def:coherence}
A path interpretability loss $\mathcal{L}$ is \emph{coherent} if the following conditions hold for any two interpretable paths $\bm{m}, \bm{m'}\in\mathcal{P}$ with respective cost sequences $\bm{c}$ and $\bm{c'}$.
\begin{enumerate}[(a)]
\item If $\bm{c}=\bm{c'}$, then $\mathcal{L}(\bm{m})=\mathcal{L}(\bm{m'})$\label{def:coherent-costdependent}.
\item (Weak Pareto dominance) If $c_k\le c'_k~\forall k$ (which we write as $\bm{c}\le\bm{c'}$), then $\mathcal{L}(\bm{m})\le\mathcal{L}(\bm{m'})$\label{def:coherent-pareto}.
\end{enumerate}
\end{definition}

Condition~\ref{def:coherent-costdependent} means that the interpretability of a path depends only on the sequence of costs along that path.
Condition~\ref{def:coherent-pareto} formalizes the intuition described before, that paths with fewer steps or better steps are more interpretable.
For instance, if we improve the cost of one step of a path while leaving all other steps unchanged, we can only make the path more interpretable.
Under any coherent interpretability loss $\mathcal{L}$ in Table~\ref{tab:three-decompositions}, $\bm{m}$ is more interpretable than $\bm{m'}$, but $\bm{m}$ may be more or less interpretable than $\bm{m}^{\text{greedy}}$ depending on the specific choice of coherent interpretability loss.

In addition, consider a path $\bm{m}\in\mathcal{P}_K$ and remove its last step to obtain a new path $\bm{m'}\in\mathcal{P}_{K-1}$.
This is equivalent to setting the $K$-th element of the cost sequence $\bm{c}(\bm{m})$ to zero.
Since $c(\cdot)> 0$, we have that $\bm{c}(\bm{m'})\le\bm{c}(\bm{m})$, which implies $\mathcal{L}(\bm{m'})\le\mathcal{L}(\bm{m})$.
In other words, under a coherent interpretability loss, removing a step from an interpretable path can only make the path more interpretable.

\begin{remark}
The path complexity $\mathcal{L}_{\text{complexity}}(\bm{m})=|\bm{m}|$ is a coherent path interpretability loss.
\end{remark}

\begin{proof}
If $\bm{m}$ and $\bm{m'}$ verify $\bm{c}(\bm{m})=\bm{c}(\bm{m'})$, then trivially the two cost sequences become zero after the same number of steps, so $\mathcal{L}_{\text{complexity}}(\bm{m})=\mathcal{L}_{\text{complexity}}(\bm{m'})$.
If $\bm{c}(\bm{m})\le\bm{c}(\bm{m'})$ and $\bm{c}(\bm{m'})$ becomes zero after exactly $K$ steps, then $\bm{c}(\bm{m})$ must become zero after at most $K$ steps, so $\mathcal{L}_{\text{complexity}}(\bm{m})\le\mathcal{L}_{\text{complexity}}(\bm{m'})$.
\end{proof}

\subsection{A Coherent Model Interpretability Loss}

Axiom~\ref{def:coherent-pareto} of Definition~\ref{def:coherence} states that a path that dominates another path in terms of the costs of each step must be at least as interpretable.
This notion of weak Pareto dominance suggests a natural path interpretability loss:
\[
\mathcal{L}_{\bm{\alpha}}(\bm{m})=\bm{\alpha}\cdot \bm{c}(\bm{m}) = \sum_{k=1}^{|\bm{m}|}\alpha_k c(m_k).
\]
In other words, the interpretability loss $\mathcal{L}_{\gamma}$ of a path $\bm{m}$ is the weighted sum of the costs of all steps in the path.
This loss function is trivially coherent and extremely general.
It is parametrized by the infinite sequence of weights $\bm{\alpha}=(\alpha_1,\alpha_2, \cdots)$, which specifies the relative importance of the accuracy of each step in the model for the particular application at hand.

Defining a family of interpretability losses with infinitely many parameters allows for significant modeling flexibility, but it is also cumbersome and overly general.
We therefore propose to select $\alpha_k=\gamma^k$ for all $k$, replacing the infinite sequence of parameters $(\alpha_1, \alpha_2, \ldots)$ with a single parameter $\gamma>0$.
In this case, following \eqref{eq:path-to-model-interpretability}, we propose the following coherent interpretability loss function on the space of models.

\begin{definition}[Model interpretability]\label{def:model-interpretability}
    Given a model $m\in\mathcal{M}$, its interpretability loss $\mathcal{L}_{\gamma}(m)$ is given by
    \begin{equation}
        \mathcal{L}_{\gamma}(m) = \begin{cases}
            \infty, &\text{if } \mathcal{P}(m) = \emptyset,\\
            \min\limits_{\bm{m} \in \mathcal{P}(m)} \mathcal{L}_{\gamma}(\bm{m})=\sum\limits_{k=1}^{|\bm{m}|}\gamma^kc(m_k), &\text{otherwise.}
        \end{cases}
    \end{equation}
\end{definition}

By definition, $\mathcal{L}_{\gamma}$ is a coherent interpretability loss, which favors more incremental models or models with a low complexity.
The parameter $\gamma$ captures the tradeoff between these two aspects of interpretability.
Theorem~\ref{thm:model-gammainfinite} shows that with a particular choice of $\gamma$ one can recover the notion of model complexity introduced in Section~\ref{sec:paths-complexity}, or models that can be built in a greedy way.

\begin{theorem}[Consistency of interpretability measure]\label{thm:model-gammainfinite}
    Assume that the cost $c(\cdot)$ is bounded, we consider $\mathcal{L}_{\gamma}(m)$ in the two limit cases $\gamma \rightarrow +\infty$ and $\gamma \rightarrow 0$:
    \begin{enumerate}[(a)]
        \item \label{enum:model-thm3} Let $m^+,~m^- \in \mathcal{M}$ with $\mathcal{L}_{\mathrm{complexity}}(m^+) < \mathcal{L}_{\mathrm{complexity}}(m^-)$ (i.e., $m^+$ requires less interpretable steps than $m^-$), or $\mathcal{L}_{\mathrm{complexity}}(m^+) = \mathcal{L}_{\mathrm{complexity}}(m^-)$ and $c(m^+) < c(m^-)$.
        \begin{equation}
            \lim_{\gamma \rightarrow \infty} \mathcal{L}_{\gamma}(m^-) - \mathcal{L}_{\gamma}(m^+) = +\infty.
        \end{equation}.
        \item \label{enum:model-thm4} 
        Given $\bm{m^+}, \bm{m^-} \in \mathcal{P}$, if $\bm{c}(\bm{m^+}) \preceq \bm{c}(\bm{m^-})$, where $\preceq$ represents the lexicographic order on $\mathbb{R}^\mathbb{N}$, then
        \begin{equation}\label{eq:model-thm10}
            \lim_{\gamma \rightarrow 0} \mathcal{L}_{\gamma}(\bm{m^-}) - \mathcal{L}_{\gamma}(\bm{m^+}) \geq 0.
        \end{equation}
        Consequently, given models $m^+, m^- \in \mathcal{M}$, if there is $\bm{m^+} \in \mathcal{P}(m^+)$ such that $\bm{c}(\bm{m^+}) \preceq \bm{c}(\bm{m^-})$ for all $\bm{m^-} \in \mathcal{P}(m^-)$, then
        \begin{equation}
            \lim_{\gamma \rightarrow 0} \mathcal{L}_{\gamma}(m^-) - \mathcal{L}_{\gamma}(m^+) \geq 0.
        \end{equation}
    \end{enumerate}
\end{theorem}

Intuitively, in the limit $\gamma \rightarrow +\infty$, \ref{enum:model-thm3} states that the most interpretable models are the ones with minimal complexity, or minimal costs if their complexity is the same.
\ref{enum:model-thm4} states that in the limit $\gamma \rightarrow 0$ the most interpretable models can be constructed with greedy steps.
Definition~\ref{def:model-interpretability} therefore generalizes existing approaches and provides a good framework to model the tradeoffs of interpretability.


%% file: 5-computation.tex

Defining an interpretability loss brings a new perspective to the literature on interpretability in machine learning.
In this section, we discuss the applications of this framework.
For the sake of generality, in the early part of this section we work with the more general interpretability loss $\mathcal{L}_{\bm{\alpha}}(\cdot)$.

\subsection{The Price of Interpretability}\label{sec:priceofinterpretability}

Given the metric of interpretability defined above, we can quantitatively discuss the price of interpretability, i.e., the tradeoff between a model's interpretability loss $\mathcal{L}_{\bm{\alpha}}(m)$ and its cost $c(m)$.
To evaluate this tradeoff, we want to compute models that are Pareto optimal with respect to $c(\cdot)$ and $\mathcal{L}_{\bm{\alpha}}(\cdot)$, as in \eqref{eq:path-fundamentalproblem}. 

Computing these Pareto-optimal solutions can be challenging, as our definition of model interpretability requires optimizing over paths of any length.
Fortunately, the only optimization problem we need to be able to solve is to find the most interpretable path of a fixed length $K$, i.e.,

\begin{equation}\label{eq:fixedpathopt}
    \min_{\bm{m} \in \mathcal{P}_K} \mathcal{L}_{\bm{\alpha}}(\bm{m}) = \sum_{k=1}^K \alpha_k c(m_k)
\end{equation}

Indeed, the following proposition shows that we can compute Pareto-optimal solutions by solving a sequence of optimization problems \eqref{eq:fixedpathopt} for various $K$ and $\bm{\alpha}$.

\begin{proposition}[Price of interpretability]\label{prop:model-priceofinterpretability}
    Pareto-optimal models that minimize the interpretability loss $\mathcal{L}_{\bm{\alpha}}$ and the cost $c(\cdot)$ can be computed by solving the following optimization problem:
    \begin{equation} \label{eq:minimize-over-k}
        \min_{K \geq 0} \left( \min_{\bf{m} \in \mathcal{P}_K} c(m_K) + \lambda \sum_{k=1}^K \alpha_k c(m_k) \right) ,
    \end{equation}
    where $\lambda \in \mathcal{R}$ is a tradeoff parameter between cost and interpretability. 
\end{proposition}

The (simple) proof of the proposition is provided in the appendix. Notice that the inner minimization problem in~\eqref{eq:minimize-over-k} is simply problem~\eqref{eq:fixedpathopt} with the modified coefficients $(\lambda\alpha_1, \ldots, \lambda\alpha_{K-1}, (1+\lambda)\alpha_K)$.

By defining the general framework of coordinate paths and a natural family of coherent interpretability loss functions, we can understand exactly how much we gain or lose in terms of accuracy when we choose a more or less interpretable model.
Our framework thus provides a principled way to answer a central question of the growing literature on interpretability in machine learning.

Readers will notice that the weighted sum of the objectives optimized in Proposition~\ref{prop:model-priceofinterpretability} does not necessarily recover the entire Pareto front, and in particular cannot recover any non-convex parts \cite{Kim2005}.

\begin{figure}
    \centering
    \includegraphics[width=0.8\columnwidth]{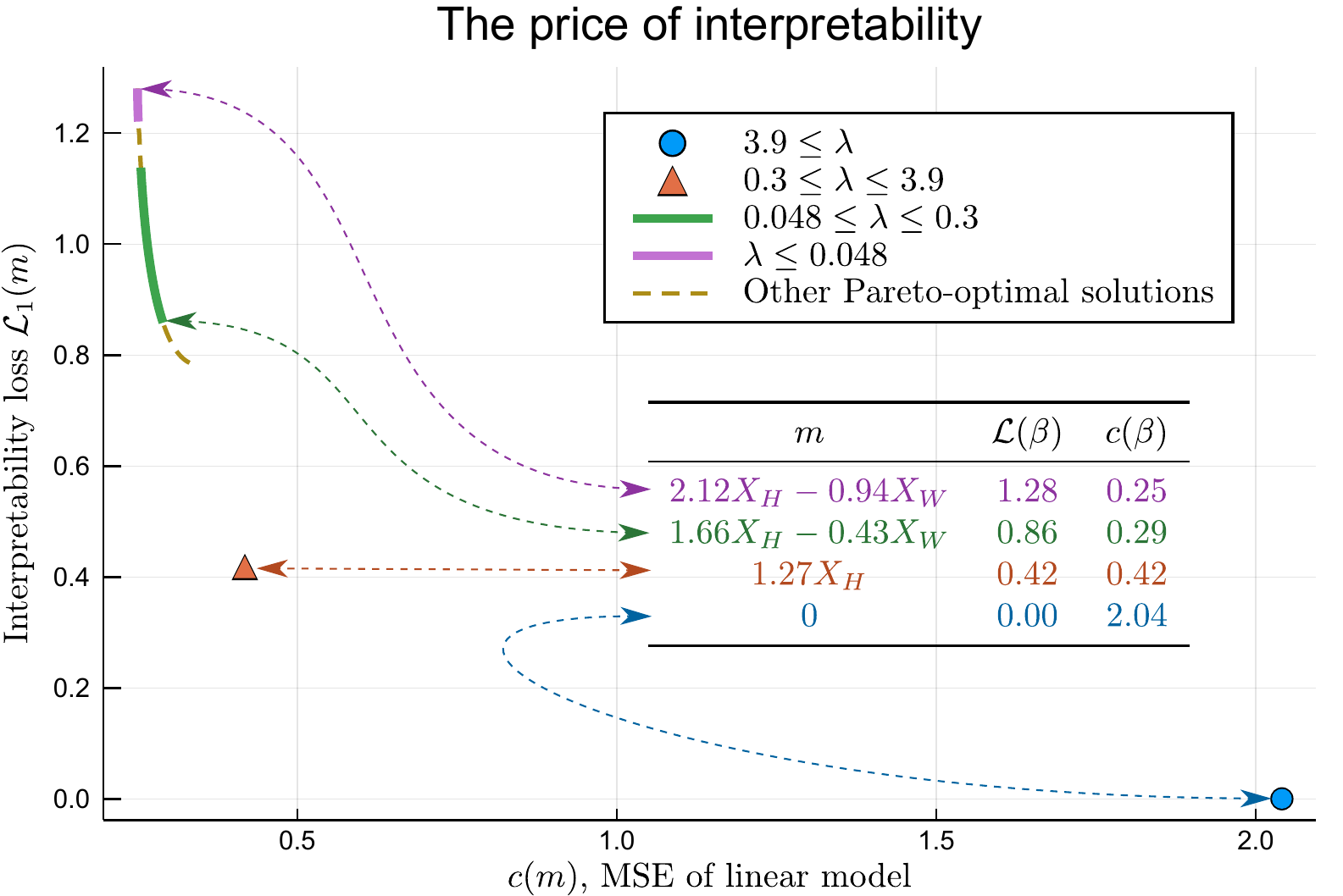}
    \caption{Pareto front between interpretability loss $\mathcal{L}(m)=\mathcal{L}_{\gamma}(m)$ (with $\gamma=1$) and cost $c(m)$ on the toy OLS problem \eqref{eq:model-linearexample}, computed by varying $\lambda$ in \eqref{eq:minimize-over-k}.
    The dashed line represents Pareto-optimal solutions that cannot be computed by this weighted-sum method.
    Note that the front is discontinuous, and that there is an infinite number of Pareto-optimal models with two steps, but only one respectively with one and zero steps.
    The inset table describes several interesting Pareto-optimal models.}
    \label{fig:models-priceofinterpretability}
\end{figure}

\begin{figure}
    \centering
    \begin{subfigure}[t]{.48\textwidth}
      \centering
      \includegraphics[height=2in]{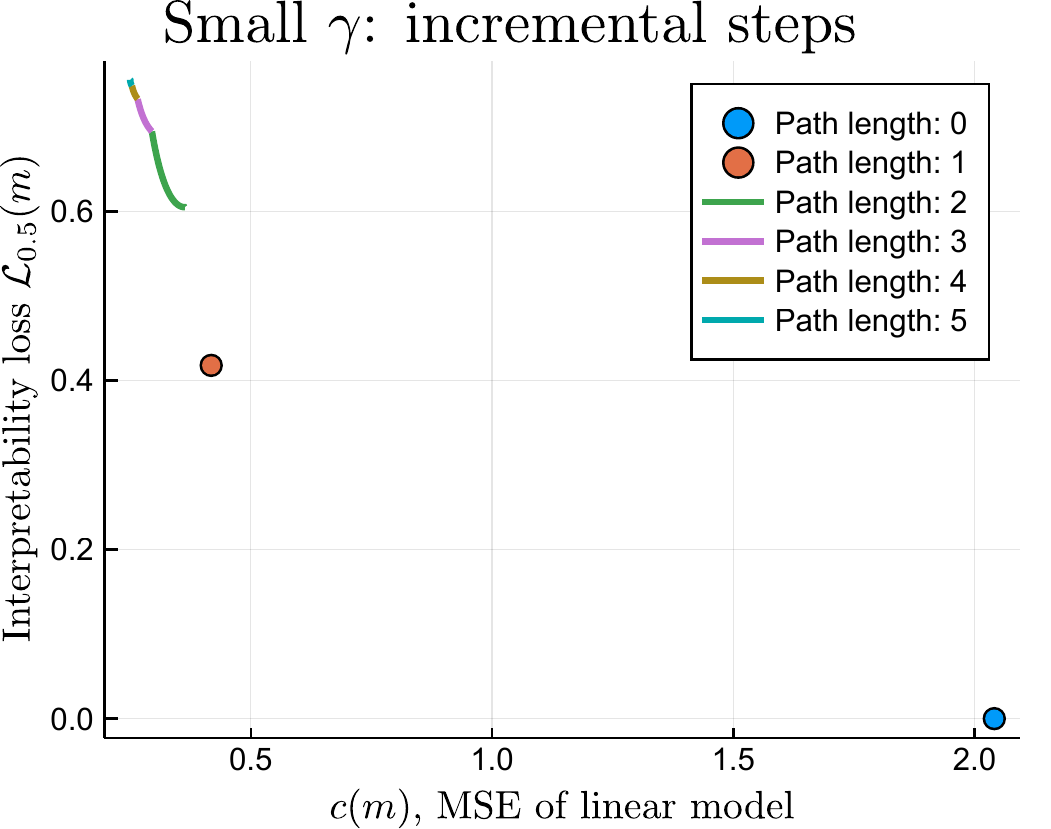}
      \caption{Here we choose $\gamma = 0.5$, therefore the first steps are the most important, and we favor incremental/greedy models, with potentially many steps (see second part of Theorem~\ref{thm:model-gammainfinite}). Each color corresponds roughly to the addition of a greedy step (this becomes exact when $\gamma \rightarrow 0$).}
      \label{fig:models-priceofinterpretability-smallgamma}
    \end{subfigure}\hfill
    \begin{subfigure}[t]{.48\textwidth}
      \centering
      \includegraphics[height=2in]{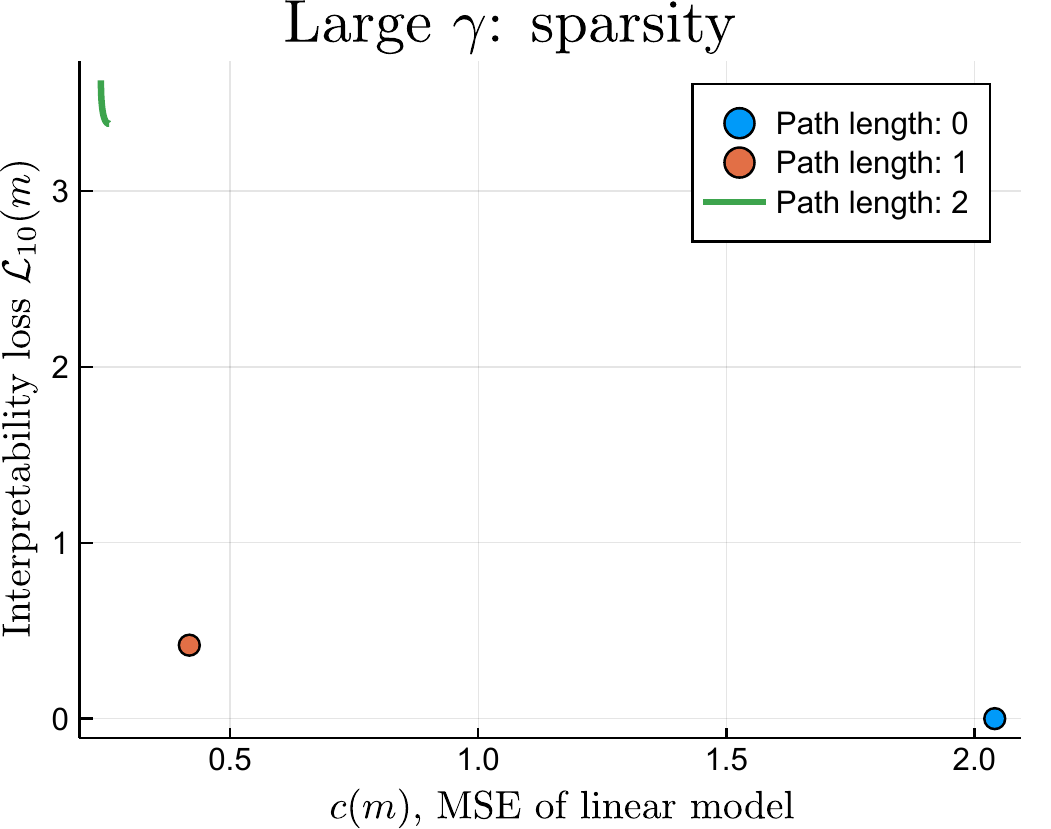}
      \caption{Here we choose $\gamma = 10$, therefore the cost of the intermediate steps is much less important than the final models, and we favor sparse models (see first part of Theorem~\ref{thm:model-gammainfinite}).
      Note that the Pareto front has three almost discrete parts, corresponding to the three possible levels of sparsity in this example.}
      \label{fig:models-priceofinterpretability-largergamma}
    \end{subfigure}
    \caption{Pareto fronts between model interpretability and cost in the same setting as Figure~\ref{fig:models-priceofinterpretability}, except that we change the definition of interpretability by changing the value $\gamma$.}
    \label{fig:models-priceofinterpretability-variousgamma}
\end{figure}

Using Proposition~\ref{prop:model-priceofinterpretability}, we can compute the price of interpretability for a range of models and interpretability losses. 
As an example, Figure~\ref{fig:models-priceofinterpretability} shows all Pareto-optimal models with respect to performance cost and interpretability for our toy problem from Section~\ref{sec:paths-tradeoff}, with the interpretability loss $\mathcal{L}_{\gamma}$ chosen such that $\gamma=1$. 
Figure~\ref{fig:models-priceofinterpretability-variousgamma} shows the Pareto front in the same setting for other values of $\gamma$.
We notice that as in Theorem~\ref{thm:model-gammainfinite}, when $\gamma$ grows large our notion of interpretability reduces to sparsity (discrete Pareto curve), whereas when $\gamma$ grows small our notion of interpretability favors a larger number of incremental steps.

\subsection{Computational Considerations}

To solve~\eqref{eq:minimize-over-k} we consider a sequence of problems of type~\eqref{eq:fixedpathopt}.
However, this sequence is possibly infinite, which poses a computational problem.
Proposition~\ref{prop:K-upperbound} provides a bound for the number of problems of type~\eqref{eq:fixedpathopt} we need to consider in the general case.

\begin{proposition} \label{prop:K-upperbound}
Assume there exist $c_{\min}$ and $c_{\max}$ such that $0 < c_{\min} \le c(m) \le c_{\max}$ for all $m\in\mathcal{M}$ (positive and bounded cost function), and consider the interpretability loss $\mathcal{L}_{\gamma}$.
If $\gamma\ge 1$, then
\begin{equation}
K_{\text{opt}}:=\arg\min_{K \geq 0} \left( \min_{\bf{m} \in \mathcal{P}_K} c(m_K) + \lambda \sum_{k=1}^K \gamma^k c(m_k) \right)\le K_{\max},
\end{equation}
where
\begin{equation} \label{eq:kmax-cases}
K_{\max}=\begin{cases}
\frac{c_{\max}}{\lambda c_{\min}} & \text{if}~\gamma = 1,\\
\frac{\log\left(1+\frac{(\gamma - 1)c_{\max}}{\lambda\gamma c_{\min}}\right)}{\log \gamma} & \text{if}~\gamma > 1.
\end{cases}
\end{equation}
\end{proposition}

In other words, under the interpretability loss $\mathcal{L}_\gamma$ with $\gamma\ge 1$, we can find the optimal solution of ~\eqref{eq:minimize-over-k} by solving at most $K_{\max}$ problems of type~\eqref{eq:fixedpathopt}.
The proof of Proposition~\ref{prop:K-upperbound} is provided in the appendix.

A corollary of Proposition~\ref{prop:K-upperbound} is that we can write an optimization formulation of problem~\eqref{eq:minimize-over-k} with a finite number of decision variables.
For instance, we can formulate the inner minimization problem with finitely many decision variables for each $K$ and then solve finitely many such problems.
The tractability of this optimization problem is application-dependent.

For example, by adapting the mixed-integer optimization formulation from Bertsimas and Dunn \cite{Bertsimas2017b}, we can compute the price of interpretability for decision trees of bounded depth by writing the following mixed-integer formulation of the inner minimization problem in~\eqref{eq:minimize-over-k}:

\begin{subequations} \label{eq:tree-formulation}
\begin{align}
\min\quad & \sum_{k=1}^K\gamma^k f(d_t^k, a_t^k, b_t^k) \\
\text{s.t.}\quad & (d_t^k, a_t^k, b_t^k) \in \mathcal{T} & \forall k\in[k] \label{eq:tree-summary}\\
&\sum_{t\in\mathcal{T}_L}d_t^k=k & \forall k\in[K] \label{eq:tree-k-splits}\\
&a_t^k\le a_t^{k+1}&\forall t\in\mathcal{T}_B, k\in[K-1]\label{eq:tree-keepsplit}\\
&d_t^k \le d_t^{k+1}&\forall t\in\mathcal{T}_B, k\in[K-1]\label{eq:tree-keepsplit2}\\
&b_t^k-(1-d_t^k) \le b_t^{k+1}\le b_t^k+(1-d_t^k)&\forall t\in\mathcal{T}_B, k\in[K-1]\label{eq:tree-splitvalue},
\end{align}
\end{subequations}
where the variables $d_t^k$, $a_t^k$ and $b_t^k$ define $K$ trees of depth at most $D$, and constraints~\eqref{eq:tree-k-splits}-\eqref{eq:tree-splitvalue} impose an interpretable path structure on the $K$ trees.
The set $\mathcal{T}_B$ indicates the set of branching nodes of the trees, the variable $d_t^k$ indicates whether branching node $t$ in tree $k$ is active, $a_t^k$ selects the variable along which to perform the split at branching node $t$ in tree $k$, and $b_t^k$ is the split value at branching node $t$ in tree $k$.
The function $f$ is the objective value of the tree defined by these split variables, and the set $\mathcal{T}$ designates all the constraints to impose the tree structure for each $k$ (constraint~\eqref{eq:tree-summary} is equivalent to (24) from \cite{Bertsimas2017b}).
Constraint~\eqref{eq:tree-k-splits} imposes that tree $k$ must have exactly $k$ active splits, Constraint~\eqref{eq:tree-keepsplit2} forces tree $k+1$ to keep all the branching nodes of tree $k$, and constraints~\eqref{eq:tree-keepsplit} and \eqref{eq:tree-splitvalue} force the splits at these common branching nodes to be the same.

\begin{figure}
\centering
\includegraphics[width=0.5\columnwidth]{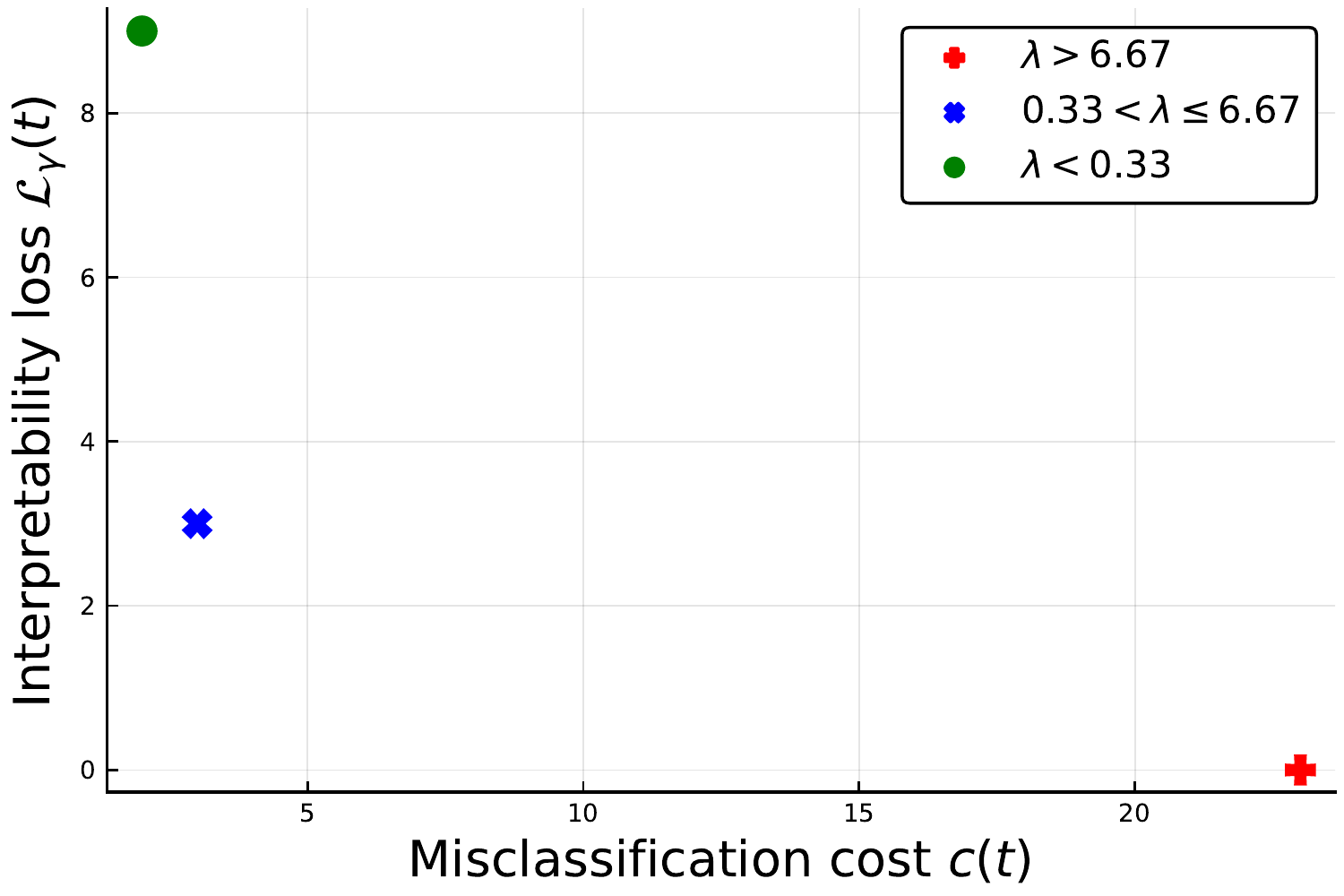}
\caption{Price of interpretability for decision trees of depth at most 2 on the simplified \texttt{iris} dataset.}
\label{fig:price-tree}
\end{figure}

This formulation allows us to compute the price of interpretability on the simplified \texttt{iris} dataset from Section~\ref{sec:paths}.
The resulting Pareto curve is shown in Figure~\ref{fig:price-tree}.
It turns out the most interpretable tree with a misclassification error of 2 is $t_{\text{good}}$, with $\mathcal{L}_\gamma(t_{\text{good}})=9$ (for $\gamma=1$).

In general, mixed-integer optimization formulations such as~\eqref{eq:tree-formulation} may not scale.
However, in many cases a provably optimal solution is not necessary and scalable heuristics such as local improvement may be employed.
We provide such an example in Section~\ref{sec:linreg}.

\subsection{Interpretable Paths and Human-in-the-Loop Analytics}

Motivated by the idea that humans read and explain models sequentially, we have used the framework of interpretable paths to evaluate the interpretability of individual models.
Viewing an interpretable path as a nested sequence of models of increasing complexity can also be useful in the context of human-in-the-loop analytics.

Consider the problem of customer segmentation via clustering.
Choosing the number of customer types ($k$) is not always obvious in practice and has to be selected by a decision-maker. Solving the clustering problem with $k$ clusters and with $k+1$ clusters may lead to very different clusters.
Alternatively, using interpretable steps, we can force a hierarchical structure on the clusters, i.e., the solution with $k+1$ clusters results from the splitting of one of the clusters of the solution with $k$ clusters, for all $k$.
The change between $k$ clusters and $k+1$ clusters becomes simpler and may facilitate the choice of $k$.

If we assume each $k$ can be chosen with equal probability for $k\leq 10$, the problem of finding the sequence that minimizes the expected cost is:
\begin{equation}
    \min_{\bm{m} \in \mathcal{P}_{10}}  \quad \frac{1}{10}\sum_{k=1}^{10} \, c(m_k),
\end{equation}
which is exactly the decision problem \eqref{eq:fixedpathopt} with the weights $\alpha_k=0.1$ for $k\leq 10$, and $\alpha_k=0$ otherwise.
This problem is related to studies in incremental approximation algorithms \cite{Lin2010} and prioritization \cite{Koc2014}, which are typically motivated by a notion of interpretability which simplifies implementation for practitioners.

More generally, we can use interpretable paths to facilitate human-in-the-loop model selection.
Given a discrete distribution on the choice of $K$ : $p_k = \mathbb{P}(\{ k \text{ will be chosen by the decision maker}\})$, we can choose $\alpha_k = p_k$ and solve \eqref{eq:fixedpathopt} to find paths $\bm{m}$ that minimize the expected cost $\mathbb{E}_k[c(m_k)]$.

%% file: 6-linreg.tex

So far, we have presented a mathematical framework to formalize the discussion of interpretability. We now study in detail how it can be used in practice, focusing on the single application of linear regression.

\subsection{Modeling interpretability}

In the example of linear regression, we defined the following interpretable steps:
\begin{equation}
\mathcal{S}(\beta)=\{\beta'\in\mathbb{R}^d : \norm{\beta-\beta'}_0\le 1\}.
\label{eq:step}
\end{equation}

These steps are a modeling choice.
They lead to decompositions of linear models where the coefficients are introduced or modified one at a time, and we have seen they are intimately linked to sparsity. 
We wish to obtain models that can easily be introduced coefficient by coefficient, allowing ourselves to modify coefficients that have already been set.

Choosing a different step function $\mathcal{S}(\cdot)$ can lead to other notions of interpretability.
For instance, each step could add a feature, allowing to modify all the weights (not only one coordinate): $\mathcal{S}_{\mathrm{features}}(\beta) = \{\theta \text{ s.t. } \|\theta\|_0 \leq \|\beta\|_0 + 1\}$.
This boils down to ordering the features of a linear model, finding the most interpretable order.
We could also choose $\mathcal{S}_\mathrm{SLIM}(\beta)=\{\theta: \norm{\beta-\theta}_0\le1, \norm{\beta-\theta}_1\in\mathbb{Z}\}$, which imposes integer coordinate updates at each step.
This is related to the notion of interpretability introduced by score-based methods \cite{Ustun2016}.
Another way to think about score-based methods is to choose $\mathcal{S}'_\mathrm{SLIM}(\beta)=\{\theta: \norm{\beta-\theta}_0\le1, \norm{\beta-\theta}_1\in\{0,1\}\}$, which imposes that each step adds one point to the scoring system.

We select the interpretability loss $\mathcal{L}_{\gamma}$ with $\gamma=1$ (meaning the costs of all steps matter equally).
Given the step function $\mathcal{S}$ defined in \eqref{eq:step}, the convex quadratic cost function $c(\cdot)$, and the initial regression coefficients $\beta_0$, as in Section~\ref{sec:computation}, our goal is to find the optimal interpretable path of length $K$ \eqref{eq:fixedpathopt}.

\subsection{Algorithms}

\paragraph{Optimal.}
Problem~\eqref{eq:fixedpathopt} can be written as a convex integer optimization problem using special ordered sets of type 1 (SOS-1 constraints).
\begin{subequations}\label{eq:linreg}
\begin{align}
\min_{\bm{\beta_k}}\quad & \sum_{k=1}^Kc(\beta_k)\\
\text{s.t.} \quad & \text{SOS-1}(\beta_{k+1}-\beta_{k}) & 0\le k < K.
\end{align}
\end{subequations}
For reasonable problem sizes ($d \leq 10$, $K \leq 10$ and any choice of $n$), this problem can be solved exactly using a standard solver such as Gurobi or CPLEX.

\paragraph{Local improvement.}
In higher-dimensional settings, or when $K$ is too large, the formulation above may no longer scale.
Thus it is of interest to develop a fast heuristic for such instances.

A feasible solution $\bm{\beta} = (\beta_1, \cdots, \beta_K)$ to problem~\eqref{eq:linreg} can be written as a vector of indices $\bm{i} = (i_1, \cdots, i_K) \in\{1,\ldots,d\}^K$ and a vector of values $\bm{\delta} = (\delta_1,\cdots, \delta_K) \in \mathbb{R}^K$, such that for $0\le k < K$,
\[
(\beta_{k+1})_i = \begin{cases}(\beta_k)_i + \delta_k, & \text{if }i = i_k\\
(\beta_k)_i, & \text{if }i\neq i_k.
\end{cases}
\]

The vector of indices $\bm{i}$ encodes which regression coefficients are modified at each step in the interpretable path, while the sequence of values $\bm{\delta}$ encodes the value of each modified regression coefficient.
Thus problem \eqref{eq:linreg} can be rewritten as
\begin{equation}
\min_{\bm{i}} \min_{\bm{\delta}}C(\bm{i}, \bm{\delta}), \qquad \text{ with } C(\bm{i}, \bm{\delta}):= \sum_{k=1}^K c(\beta_0 + \sum_{j=1}^k\delta_j e_{i_j}),
\end{equation}
where $e_i$ designates the $i$-th unit vector.
Notice that the inner minimization problem is an ``easy'' convex quadratic optimization problem, while the outer minimization problem is a ``hard'' combinatorial optimization problem.
We propose the following local improvement heuristic for the outer problem: given a first sequence of indices $\bm{i} = \bm{i}^0$, we randomly sample one step $\kappa$ in the interpretable path.
Keeping all $i_k$ constant for $k\neq\kappa$, we iterate through all $d$ possible values of $i_{\kappa}$ and obtain $d$ candidate vectors $\bm{\hat{i}}$.
For each candidate, we solve the inner minimization problem and keep the one with lowest cost.
The method is described in full detail as Algorithm~\ref{algo:local-improvement}, in the more general case where we sample not one but $q$ steps from the interpretable path.

\begin{algorithm}
\caption{Local improvement heuristic.
Inputs: regression cost function $c(\cdot)$; starting vector of indices $i^0$.
Parameters: $q\in\mathbb{N}$ controls the size of the neighborhood, $T\in\mathbb{N}$ controls the number of iterations.}\label{algo:local-improvement}
\begin{algorithmic}[1]
\Function{LocalImprovement}{$c(\cdot)$, $\bm{i}^0$, $q$, $T$}
\For{$1 \le t \le T$}
\State $\bm{i}^* \gets \bm{i}^0$
\State $\bm{\delta}^* \gets \arg\min_{\bm{\delta}} C\left(\bm{i}^0, \bm{\delta}\right)$
\State $C^* \gets C\left(\bm{i}^0, \bm{\delta}^*\right)$
\State Randomly select $\mathcal{K}=\{\kappa_1, \ldots, \kappa_q\} \subset \{1,\ldots,K\}$ \Comment{subset of cardinality $q$}
\State $\bm{\hat{i}} \gets \bm{i}^*$
\State $\bm{\hat{\delta}} \gets \bm{\delta}^*$
\For{$(f_1, \ldots, f_q) \in \{1, \ldots, d\}^q$}
\For{$1 \le p \le q$}
\State $\hat{i}_{\kappa_p} = f_p$
\EndFor
\State $\bm{\hat{\delta}} \gets \arg\min_{\bm{\delta}} C\left(\bm{\hat{i}}, \bm{\delta}\right)$
\If{$C\left(\bm{\hat{i}}, \bm{\hat{\delta}}\right) < C^*$}
\State $C^* \gets C\left(\bm{\hat{i}}, \bm{\hat{\delta}}\right)$
\State $\bm{i}^* \gets \bm{\hat{i}}$
\State $\bm{\delta}^* \gets \hat{\delta}$
\EndIf
\EndFor
\EndFor
\State \textbf{return} $\bm{i}^*, \bm{\delta}^*$
\EndFunction
\end{algorithmic}
\end{algorithm}

In order to empirically evaluate the local improvement heuristic, we run it with different batch sizes $q$ on a small real dataset, with 100 rows and 6 features (after one-hot encoding of categorical features).
The goal is to predict the perceived prestige (from a survey) of a job occupation given features about it, including education level, salary, etc.

Given this dataset, we first compute the optimal coordinate path of length $K=10$.
We then test our local improvement heuristic on the same dataset.
Given the small size of the problem, in the complete formulation a provable global optimum is found by Gurobi in about 5 seconds.
To be useful, we would like our local improvement heuristic to find a good solution significantly faster.
We show convergence results of the heuristic for different values of the batch size parameter $q$ in Table~\ref{tab:convergence}. 
For both batch sizes, the local improvement heuristic converges two orders of magnitude faster than Gurobi.
With a batch size $q=2$, the solution found is optimal.

\subsubsection{Results}

\begin{figure}
\begin{subfigure}[c]{0.37\columnwidth}
\centering
\scriptsize
\centering
\begin{tabular}{lr}
\toprule
Feature name & Description\\
\midrule
Enrollment & Total enrollment\\
Teachers & Number of teachers\\
CalWPct & \% receiving state aid\\
MealPct & \% with subsidized lunch\\
Computers & Number of computers\\
CompStu & Computers per student\\
ExpnStu & Expenditure per student\\
StuTeach & Student-teacher ratio\\
AvgInc & Average income (district)\\
ELPct & \% English Learners\\
\bottomrule
\end{tabular}
\caption{Features of the test score dataset.}
\label{tab:features}
\vspace{2ex}
\begingroup
\footnotesize
\begin{tabular}{lcccc|r}
\toprule
& \multicolumn{4}{c}{Feature}\\
Step & MealPct & AvgInc & ELPct & ExpnStu & MSE\\
\midrule
0 & $-0.87$ & - & - & - & 0.122\\
1 & $-0.87$ & $\bm{0.23}$ & - & - & 0.122\\
2 & $\bm{-0.59}$ & 0.23 & - & - & 0.117\\
3 & $-0.59$ & 0.23 & $\bm{-0.18}$ & - & 0.099\\
4 & $-0.59$ & 0.23 & $-0.18$ & $\bm{0.07}$ & 0.097\\
\bottomrule
\end{tabular}
\endgroup
\caption{Path from old model to new model.}
\label{fig:path-caschool-details}
\end{subfigure}\hfill
\begin{subfigure}[c]{0.62\columnwidth}
\includegraphics[width=\columnwidth]{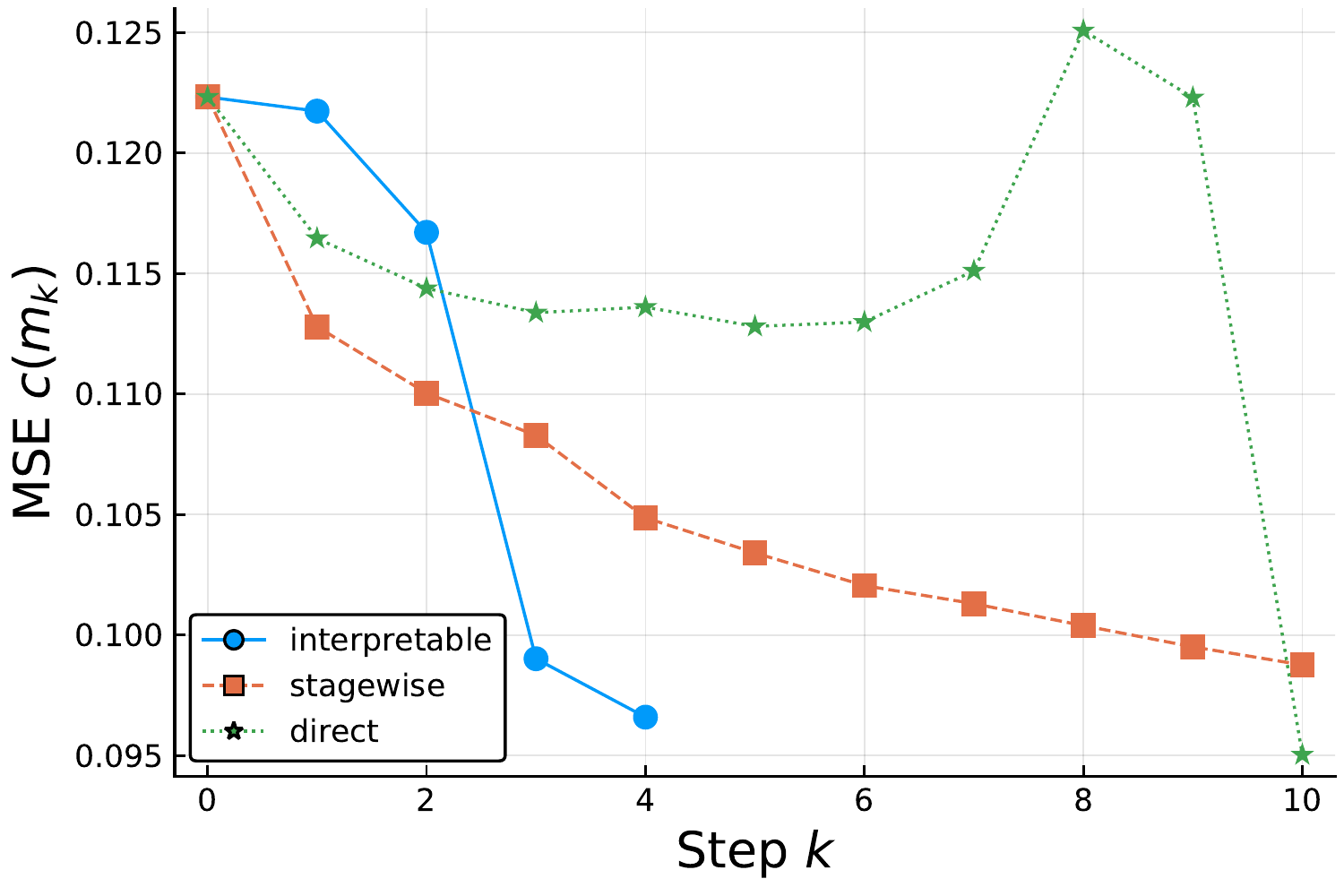}
\caption{Comparison between interpretable path and other approaches}
\label{fig:path-caschool-comparison}
\end{subfigure}
\caption{Example of a Pareto-efficient interpretable path.
On the left we see the benefits of each coefficient modification.
On the right we compare the interpretable path with two other possible paths.
The first is the forward stagewise path which greedily selects the best $m_{k+1}$ given $m_k$.
The second is a ``direct'' path, which adds the optimal least squares coefficients one by one.
The direct method is only good when all the coefficients have been added, whereas the greedy approach is good at first but then does not converge.
The interpretable path is willing to make some steps that do not improve the cost too much in preparation for very cost-improving steps.}
\label{fig:path-caschool}
\end{figure}

We now explore the results of the presented approach on a dataset of test scores in California from 1998-1999.
Each data point represents a school, and the variable of interest is the average standardized test score of students from that school.
All features are continuous and a full list is presented in Table~\ref{tab:features}.
Both the features and the target variables are centered and rescaled to have unit variance.

In our example, we assume that we already have a regression model available to predict the averaged test score: it was trained using only the percentage of students qualifying for a reduced-price lunch.
This model has an MSE of 0.122 (compared to an optimal MSE of 0.095). We would like to update this model in an interpretable way given the availability of all features in the dataset. This corresponds to the problem of constructing an interpretable path, as before, with the simple modification that $m_0$ no longer designates the regression model $\bm{0}$, but an arbitrary starting model (in particular, the one we have been provided).

The first thing we can do is explore the price of interpretability in this setting.
We can use the method presented in Section~\ref{sec:priceofinterpretability} to compute find Pareto efficient interpretable models.
The resulting price curve is shown in Figure~\ref{fig:pareto-caschool-total}.

\begin{minipage}{0.43\columnwidth}
\begin{table}[H]
\centering
\begin{tabular}{lrr}
\toprule
Method & Time (s) & Gap (\%)\\
\midrule
Exact & $5.078$ & $0.00$\\
Local imp. ($q=1$) & $0.004$ & $0.02$\\
Local imp. ($q=2$) & $0.019$ & $0.00$\\
\bottomrule
\end{tabular}
\vspace{4mm}
\caption{Convergence time and optimality gap of local improvement heuristics for different batch sizes $q$.}
\label{tab:convergence}
\end{table}
\end{minipage}\hfill
\begin{minipage}{0.5\columnwidth}
\begin{figure}[H]
\centering
\includegraphics[width=\columnwidth]{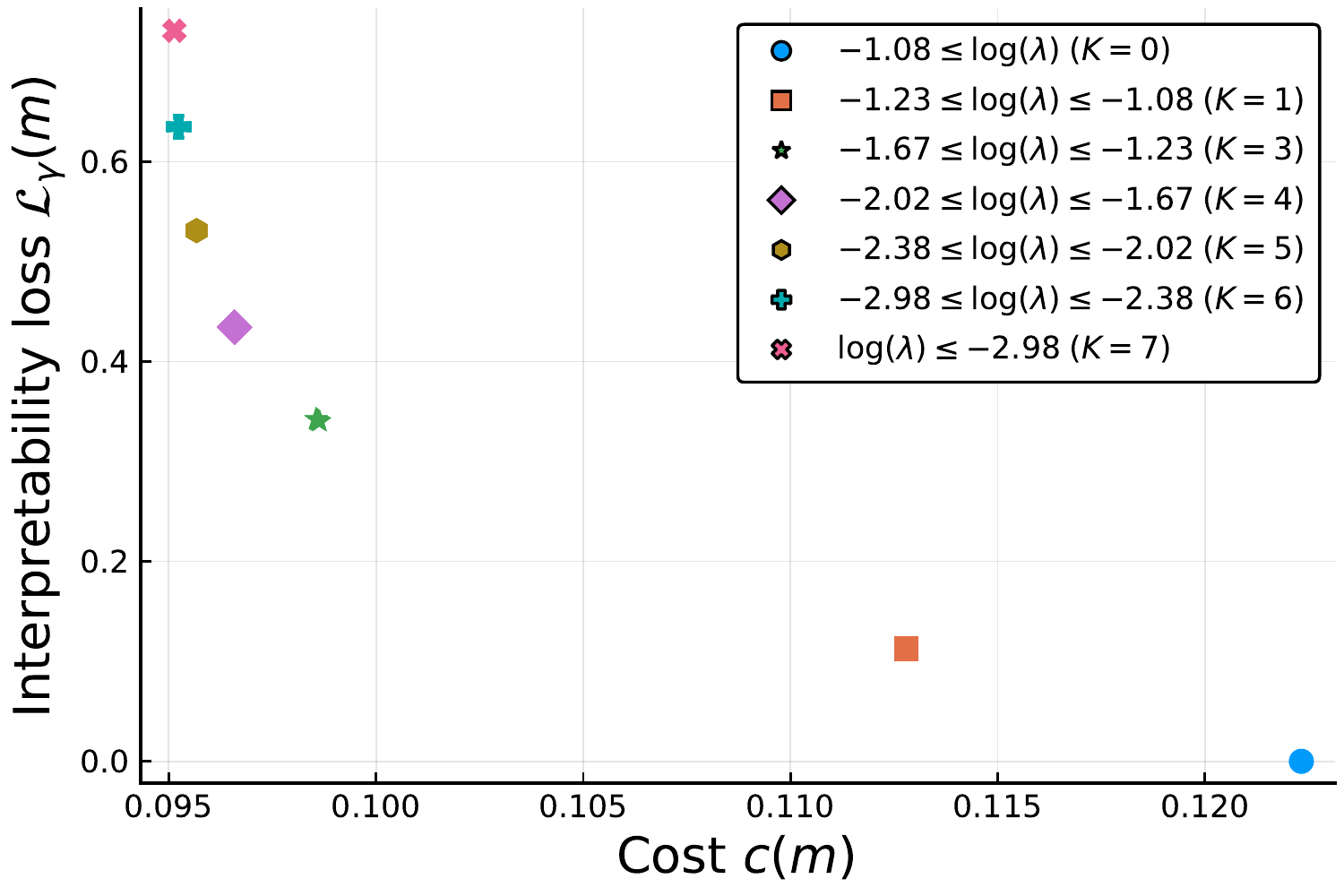}
\caption{Pareto-efficient models from the perspective of interpretability and cost.}
\label{fig:pareto-caschool-total}
\end{figure}
\vspace{5pt}
\end{minipage}

Given this price curve, we choose $\log(\lambda)\approx-1.65$ because it yields an accurate final model while avoiding diminishing interprtability returns.
This yields the new model (and associated interpretable path) shown in Figure~\ref{fig:path-caschool-details}.
This new model can be obtained from the old in just four steps.
First we add the district average income with a positive coefficient, then we correct the coefficient for reduced-price lunch students to account for this new feature, and finally we add the percentage of English learners and the school's per-student spending.
The final model has an MSE of 0.097 which is near-optimal.
When we compare this path to other methods (see Figure~\ref{fig:path-caschool-comparison}) we see that our interpretable formulation allows us to find a good tradeoff between a greedy, ``every step must improve'' formulation, and a formulation that just sets the coefficients to their final values one by one.

%% file: appendix.tex

\subsection{Proof of Theorem~\ref{thm:model-gammainfinite}}

\begin{proof}[Proof of part \ref{enum:model-thm3}]
    As $c(\cdot)$ is bounded, we have $c_{\max} \in \R$ such that $0 < c(\cdot) \leq c_{\max}$.

    Let $\bm{m^+} \in \mathcal{P}(m^+)$ be a path of optimal length to the model $m^+$, i.e., $|\bm{m^+}| = \mathcal{L}_\mathrm{complexity}(m^+)$. 
    Let $\bm{m^-} \in \mathcal{P}(m^-)$ be any path leading to $m^-$ (not necessarily of optimal length).
    By assumption, we have $|\bm{m^-}| \geq |\bm{m^+}|$, and by definition of model interpretability, we have $\mathcal{L}_\gamma(m^+) \leq \mathcal{L}_\gamma(\bm{m^+})$.
    Therefore we obtain:
    \begin{align}
        &\mathcal{L}_\gamma(\bm{m^-}) - \mathcal{L}_\gamma(m^+) \geq \mathcal{L}_\gamma(\bm{m^-}) - \mathcal{L}_\gamma(\bm{m^+}) \\
        &\quad = \sum_{k=1}^{|\bm{m^-}|} \gamma^k\, c(m^-_k) - \sum_{k=1}^{|\bm{m^+}|} \gamma^k\, c(m^+_k) \label{eq:model-proofthmconsistency-1}\\
        &\quad = \gamma^{|\bm{m^+}|} \parenth{\sum_{k=1}^{|\bm{m^+}| - 1} \frac{1}{\gamma^{|\bm{m^+}|-k}}\, \parenth{c(m^-_k) - c(m^+_k)} +  \parenth{c(m^-_{|\bm{m^+}|}) - c(m^+_{|\bm{m^+}|})} + \sum_{k=|\bm{m^+}| + 1}^{|\bm{m^-}|} \gamma^{k-|\bm{m^+}|}\, c(m^-_k) } \label{eq:model-proofthmconsistency-2}\\
        &\quad \geq \gamma^{|\bm{m^+}|} \parenth{- c_{\max} \sum_{k=1}^{|\bm{m^+}| - 1} \frac{1}{\gamma^{|\bm{m^+}|-k}} + \left( c(m^-_{|\bm{m^+}|}) - c(m^+) \right) +  \sum_{k=|\bm{m^+}| + 1}^{|\bm{m^-}|} \gamma^{k-|\bm{m^+}|}\, c(m^-_k)},\label{eq:model-proofthmconsistency-3}
    \end{align} 
    where \eqref{eq:model-proofthmconsistency-1} follows from the definition of model interpretability, \eqref{eq:model-proofthmconsistency-2} is just a development of the previous equation, and \eqref{eq:model-proofthmconsistency-3} just bounds the first sum and uses $m^+_{|\bm{m^+}|} = m^+$ for the middle term.

    If $\mathcal{L}_\mathrm{complexity}(m^+) < \mathcal{L}_\mathrm{complexity}(m^-)$, we have $|\bm{m^+}| < |\bm{m^-}|$, and therefore the last sum in \eqref{eq:model-proofthmconsistency-3} is not empty and for $\gamma \geq 1$ we can bound it:
    \begin{equation}
        \sum_{k=|\bm{m^+}| + 1}^{|\bm{m^-}|} \gamma^{k-|\bm{m^+}|}\, c(m^-_k) \geq \gamma^{|\bm{m^-}|-|\bm{m^+}|} \, c(m_{|\bm{m^-}|}^-) \geq \gamma \, c(m^-).
    \end{equation}

    Therefore, for $\gamma \geq 1$ we have:
    \begin{equation}
        \mathcal{L}_\gamma(\bm{m^-}) - \mathcal{L}_\gamma(m^+) \geq \gamma^{|\bm{m^+}|} \left( \gamma \, c(m^-) - c(m^+ ) - c_{max} \sum_{k=1}^{|\bm{m^+}| - 1} \frac{1}{\gamma^{|\bm{m^+}|-k}}\right).
    \end{equation}

    This bound is valid for all the path $\bm{m^-}$ leading to $m^-$, in particular the one with optimal interpretability loss, therefore we have (for $\gamma \geq 1$): 
    \begin{equation}
        \mathcal{L}_\gamma(m^-) - \mathcal{L}_\gamma(m^+) \geq \gamma^{|\bm{m^+}|} \left( \gamma \, c(m^-) - c(m^+ ) - c_{max} \sum_{k=1}^{|\bm{m^+}| - 1} \frac{1}{\gamma^{|\bm{m^+}|-k}}\right).
    \end{equation}
    which implies (as $c(m^-) > 0$):
    \begin{equation}
        \lim_{\gamma \rightarrow +\infty} \mathcal{L}_\gamma(m^-) - \mathcal{L}_\gamma(m^+) = +\infty
    \end{equation}

    We now look at the case  $\mathcal{L}_\mathrm{complexity}(m^+) = \mathcal{L}_\mathrm{complexity}(m^-)$ and $c(m^+) < c(m^-)$. For $\gamma \geq 1$, we can easily bound parts of equation \eqref{eq:model-proofthmconsistency-3}:
    \begin{equation}
        c(m^-_{|\bm{m^+}|}) + \sum_{k=|\bm{m^+}| + 1}^{|\bm{m^-}|} \gamma^{k-|\bm{m^+}|}\, c(m^-_k) \geq \gamma^{|\bm{m^-}|-|\bm{m^+}|} c(m^-_{|\bm{m^-}|})\geq c(m^-).
    \end{equation}

    Putting it back into \eqref{eq:model-proofthmconsistency-3}, we obtain (for $\gamma \geq 1$)
    \begin{equation}
        \mathcal{L}_\gamma(\bm{m^-}) - \mathcal{L}_\gamma(m^+) \geq \gamma^{|\bm{m^+}|} \left( \left( c(m^-) - c(m^+) \right) - c_{max} \sum_{k=1}^{|\bm{m^+}| - 1} \frac{1}{\gamma^{|\bm{m^+}|-k}}\right).
    \end{equation}

    This bound is independent of the path $\bm{m^-}$ leading to $m^-$, therefore we have 
    \begin{equation}
        \mathcal{L}_\gamma(m^-) - \mathcal{L}_\gamma(m^+)  \geq \gamma^{|\bm{m^+}|} \left( \left( c(m^-) - c(m^+) \right) - c_{max} \sum_{k=1}^{|\bm{m^+}| - 1} \frac{1}{\gamma^{|\bm{m^+}|-k}}\right) \rightarrow_{\gamma \rightarrow +\infty} +\infty,
    \end{equation}
    which ends the proof of part~\ref{enum:model-thm3} of Theorem~\ref{thm:model-gammainfinite}.
\end{proof}

\begin{proof}[Proof of part \ref{enum:model-thm4}]
    Consider two paths $\bm{m^+}, \bm{m^-} \in \mathcal{P}$, such that $\bm{c}(\bm{m^+}) \preceq \bm{c}(\bm{m^-})$.
    By definition of the lexicographic order, either the two paths are the same (in that case the theorem is trivial), or there exist $K \geq 1$ such that:
    \begin{equation*}
        \begin{cases}
            \bm{c}(\bm{m^+})_k = \bm{c}(\bm{m^-})_k \quad \forall k < K \\
            \bm{c}(\bm{m^+})_K < \bm{c}(\bm{m^-})_K .
        \end{cases}
    \end{equation*}
    
    We have:
    \begin{align}
        \mathcal{L}_\gamma(\bm{m^-}) - \mathcal{L}_\gamma(\bm{m^+}) &= \sum_{k=1}^{|\bm{m^-}|} \gamma^k\, c(m^-_k) - \sum_{k=1}^{|\bm{m^+}|} \gamma^k\, c(m^+_k) \\
        &= \sum_{k=1}^{\infty} \gamma^k\, \left( \bm{c}(\bm{m^-})_k - \bm{c}(\bm{m^+})_k \right) \label{eq:model-proofthmconsistency-4}\\
        &= \sum_{k=1}^{K-1} \gamma^k\, \left( \bm{c}(\bm{m^-})_k - \bm{c}(\bm{m^+})_k \right) + \gamma^K\left( \bm{c}(\bm{m^-})_K - \bm{c}(\bm{m^+})_K \right) \nonumber\\
        &\qquad \quad + \sum_{k=K+1}^{\infty} \gamma^k\, \left( \bm{c}(\bm{m^-})_k - \bm{c}(\bm{m^+})_k \right)\\
        &= \gamma^K \left( \bm{c}(\bm{m^-})_K - \bm{c}(\bm{m^+})_K + \sum_{k=K+1}^{\infty} \gamma^{k-K}\, \left( \bm{c}(\bm{m^-})_k - \bm{c}(\bm{m^+})_k \right)\right), \label{eq:model-proofthmconsistency-5}
    \end{align}
    where \eqref{eq:model-proofthmconsistency-4} just applies the definition of the sequence $\bm{c}$, and \eqref{eq:model-proofthmconsistency-5} uses $\bm{c}(\bm{m^+})_k = \bm{c}(\bm{m^-})_k \quad \forall k < K$.
    
    The term inside the parenthesis in \eqref{eq:model-proofthmconsistency-5} converges to $\bm{c}(\bm{m^-})_K - \bm{c}(\bm{m^+})_K > 0$ when $\gamma \rightarrow 0$, as the paths are finite. Therefore  
    \begin{equation}
        \lim_{\gamma \rightarrow 0} \mathcal{L}_\gamma(\bm{m^-}) - \mathcal{L}_\gamma(\bm{m^+}) \geq 0,
    \end{equation}
    which proves \eqref{eq:model-thm10}.
    The very end of the theorem is an immediate consequence.
\end{proof}

\subsection{Proof of Proposition~\ref{prop:model-priceofinterpretability}}

\begin{proof}
    First, a solution of 
    \begin{equation*}
        \min_{m \in \mathcal{M}} \left(c(m) + \lambda \mathcal{L}_{\bm{\alpha}}(m)\right)
    \end{equation*}
    is Pareto optimal between the cost $c(\cdot)$ and the interpretability $\mathcal{L}_{\bm{\alpha}}(\cdot)$ as it corresponds to the minimization of a weighted sum of the objectives.
    Furthermore, we can write
    \begin{align*}
        \min_{m \in \mathcal{M}} \left(c(m) + \lambda \mathcal{L}_{\bm{\alpha}}(m)\right)
        =& \min_{m \in \mathcal{M}} \left(c(m) + \lambda \min_{\bm{m}\in \mathcal{P}(m)} \mathcal{L}_{\bm{\alpha}}(\bm{m})\right)
        = \min_{m \in \mathcal{M},\, \bm{m}\in \mathcal{P}(m)} \left(c(m) + \lambda \mathcal{L}_{\bm{\alpha}}(\bm{m})\right)\\
        =& \min_{m \in \mathcal{M},\, K \geq 0,\, \bm{m}\in \mathcal{P}_K(m)}
        \left(c(m_K) + \lambda \sum_{k=1}^K\alpha_k c(m_k)\right)\\
        =& \min_{K \geq 0,\, \bm{m}\in \mathcal{P}_K} \left(c(m_K) + \lambda \sum_{k=1}^K\alpha_k c(m_k)\right).
\end{align*}
\end{proof}

\subsection{Proof of Proposition~\ref{prop:K-upperbound}}

\begin{proof}
For any $K\ge 0$ and $\lambda>0$, define the optimal objective
\[
z_{\lambda}(K)=\min_{\bm{m} \in \mathcal{P}_K} c(m_K) + \lambda\sum_{k=1}^K \gamma^k c(m_k).
\]
Because $c(\cdot)$ is bounded below by $c_{\min}$, we can write
\begin{equation} \label{eq:proof-general-bound}
z_{\lambda}(K) \ge c_{\text{min}}+\lambda\sum_{k=1}^K \gamma^k c_{\min}\ge \lambda c_{\min}\sum_{k=1}^K \gamma^k.
\end{equation}
By definition $z_{\lambda}(0)$ is the cost of the empty model, so by the boundedness of $c(\cdot)$, we have $z_{\lambda}(0)\le c_{\max}$. Consider first the case when $\gamma=1$.
Then \eqref{eq:proof-general-bound} simplifies to
\[
z_{\lambda}(K) \ge \lambda K c_{\min}.
\]
Setting $K\ge K_{\max}:=c_{\max}/(\lambda c_{\min})$ yields $z_{\lambda}(K)\ge c_{\max} \ge Z_{\lambda}(0)$ and so the interpretable path of length $0$ has a better objective than any path of length at least $K_{\max}$. Now consider $\gamma>1$. In this case, \eqref{eq:proof-general-bound} simplifies to
\[
z_{\lambda}(K) \ge \lambda c_{\min}\gamma\frac{1-\gamma^K}{1-\gamma}.
\]
Defining $K_{\max}$ as in~\eqref{eq:kmax-cases}, we again see that the interpretable path of length $0$ has a better objective than any path of length at least $K_{\max}$, which completes the proof.
\end{proof}